\documentclass[10pt,twocolumn,letterpaper]{article}

\makeatletter
\@namedef{ver@everyshi.sty}{}
\makeatother
\usepackage{cvpr}              

\usepackage{graphicx}
\usepackage{amsmath}
\usepackage{amssymb}
\usepackage{booktabs}
\usepackage[pagebackref,breaklinks,colorlinks]{hyperref}

\usepackage{url}            
\usepackage{amsfonts}       
\usepackage{nicefrac}       
\usepackage{microtype}      
\usepackage{xcolor,xspace}         
\usepackage{color}
\usepackage{algorithm}
\usepackage{algpseudocode}
\usepackage{multirow}
\usepackage{amsthm}
\usepackage{thmtools}
\usepackage{microtype}
\usepackage{comment}
\usepackage{pifont}
\usepackage{subcaption}
\usepackage{graphicx}
\usepackage{paralist} 
\usepackage{wrapfig}
\usepackage{textcomp}
\hypersetup{
    colorlinks=true,
    linkcolor=blue,
    urlcolor=blue,
}
\usepackage{epsfig}
\usepackage{microtype}

\def\eg{\emph{e.g}\onedot} \def\ie{\emph{i.e}\onedot}

\usepackage[colorinlistoftodos,bordercolor=orange,backgroundcolor=orange!20,linecolor=orange,textsize=scriptsize]{todonotes}

\usepackage[capitalize]{cleveref}
\crefname{section}{Sec.}{Secs.}
\Crefname{section}{Section}{Sections}
\Crefname{table}{Table}{Tables}
\crefname{table}{Tab.}{Tabs.}

\def\cvprPaperID{5819} 
\def\confName{CVPR}
\def\confYear{2023}

\begin{document}

\title{Computationally Budgeted Continual Learning: What Does Matter?}
\author{Ameya Prabhu$^{1}$\thanks{authors contributed equally; order decided by a coin flip.} \qquad
Hasan Abed Al Kader Hammoud$^{2*}$ \qquad
Puneet Dokania$^{1}$ \qquad Philip H.S. Torr$^{1}$\\
Ser-Nam Lim$^{3}$ \qquad Bernard Ghanem$^{2}$ \qquad Adel Bibi$^{1}$\\
$^1$University of Oxford \qquad $^2$King Abdullah University of Science and Technology (KAUST) \qquad $^3$Meta AI\\
}

\maketitle
\begin{abstract}
Continual Learning (CL) aims to sequentially train models on streams of incoming data that vary in distribution by preserving previous knowledge while adapting to new data. Current CL literature focuses on restricted access to previously seen data,  while imposing no constraints on the computational budget for training. This is unreasonable for applications in-the-wild, where systems are primarily constrained by computational and time budgets, not storage. We revisit this problem with a large-scale benchmark and analyze the performance of traditional CL approaches in a compute-constrained setting, where effective memory samples used in training can be implicitly restricted as a consequence of limited computation. We conduct experiments evaluating various CL sampling strategies, distillation losses, and partial fine-tuning on two large-scale datasets, namely ImageNet2K and Continual Google Landmarks V2 in data incremental, class incremental, and time incremental settings. Through extensive experiments amounting to a total of over 1500 GPU-hours, we find that, under compute-constrained setting, traditional CL approaches, with no exception, fail to outperform a simple minimal baseline that samples uniformly from memory. Our conclusions are consistent in a different number of stream time steps, \eg, 20 to 200, and under several computational budgets. This suggests that most existing CL methods are particularly too computationally expensive for realistic budgeted deployment. Code for this project is available at: \url{https://github.com/drimpossible/BudgetCL}.
\vspace{-0.25cm}
\end{abstract}

\section{Introduction}
\vspace{-0.1cm}

\begin{figure}
\centering
\includegraphics[width=1\linewidth]{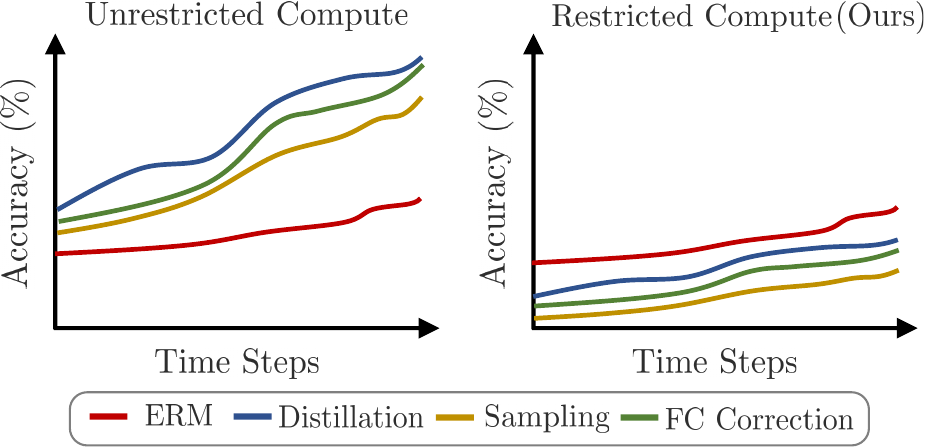}
\caption{\textbf{Main Findings. } Under per time step computationally budgeted continual learning, classical continual learning methods, \eg, sampling strategies, distillation losses, and fully connected (FC) layer correction based methods such as calibration, struggle to cope with such a setting. Most proposed continual algorithms are particularly useful only when large computation is available, where, otherwise, minimalistic algorithms (ERM) are superior.}
\label{fig:pull_figure}
\vspace{-0.60cm}
\end{figure}

Deep learning has excelled in various computer vision tasks \cite{he2016deep, schmidhuber2015deep,bommasani2021opportunities,lecun2015deep} by performing hundreds of shuffled passes through well-curated offline static labeled datasets. However, modern real-world systems, \eg, Instagram, TikTok, and Flickr, experience high throughput of a constantly changing stream of data, which poses a challenge for deep learning to cope with such a setting. Continual learning (CL) aims to go beyond static datasets and develop learning strategies that can adapt and learn from streams where data is presented incrementally over time, often referred to as time steps. However, the current CL literature overlooks a key necessity for practical real deployment of such algorithms. In particular, most prior art is focused on \textit{offline continual learning} \cite{rebuffi2017icarl,hou2019learning,kirkpatrick2017overcoming} where, despite limited access to previous stream data, training algorithms do not have restrictions on the computational training budget per time step. 

High-throughput streams, \eg, Instagram, where every stream sample at every time step needs to be classified for, say, misinformation or hate speech, are time-sensitive in which long training times before deployment are simply not an option. Otherwise, new stream data will accumulate until training is completed, causing server delays and worsening user experience. 

Moreover, limiting the computational budget is necessary towards reducing the overall cost. This is because computational costs are higher compared to any storage associated costs. For example, on Google Cloud Standard Storage (2\textcent~per GB per month), it costs no more than 6\textcent~ to store the entire CLEAR benchmark \cite{lin2021clear}, a recent large-scale CL dataset. On the contrary, one run of a CL algorithm on CLEAR performing $\sim 300$K iterations costs around $100$\$ on an A100 Google instance (3\$ per hour for 1 GPU). Therefore, it is prudent to have computationally budgeted methods where the memory size, as a consequence, is implicitly restricted. This is because, under a computational budget, it is no longer possible to revisit all previous data even if they were all stored in memory (given their low memory costs). 

This raises the question: ``\textit{Do existing continual learning algorithms perform well under per step restricted computation?"} To address this question, we exhaustively study continual learning systems, analyzing the effect of the primary directions of progress proposed in the literature in the setting where algorithms are permitted fixed computational budget per stream time step. We evaluate and benchmark at scale various classical CL sampling strategies (Uniform, Class-Balanced\cite{prabhu2020gdumb}, Recency-Biased\cite{lin2021clear}, FIFO \cite{chaudhry2019continual, cai2021online}, Max Loss, Uncertainity Loss \cite{bang2021rainbow}, and KMeans \cite{chaudhry2019continual}), CL distillation strategies (BCE \cite{rebuffi2017icarl}, MSE \cite{buzzega2020dark}, CrossEntropy \cite{wu2019large}, and Cosine \cite{hou2019learning}) and FC layer corrections (ACE \cite{zeno2018task,mai2022online,caccia2022new}, BiC \cite{wu2019large}, CosFC \cite{hou2019learning}, and WA \cite{zhao2020maintaining}) that are common in the literature. Evaluation is carried on two large-scale datasets, amounting to a total of 1500 GPU-hours, namely ImageNet \cite{deng2009imagenet} and Continual Google Landmarks V2 \cite{prabhu2023online} (CGLM) under various stream settings, namely, data incremental, class incremental, and time incremental settings. We compare against Naive; a simple baseline that, utilizing all the per step computational budget, trains while sampling from previous memory samples. 

\noindent \textbf{Conclusions.} We summarize our empirical conclusions in three folds. \textbf{(1)} None of the proposed CL algorithms, see Table \ref{fig:pull_figure} for considered methods, can outperform our simple baseline when computation is restricted. \textbf{(2)} The gap between existing CL algorithms and our baseline becomes larger with harsher compute restrictions. \textbf{(3)} We find that training a minimal subset of the model can close the performance gap compared to our baseline in our setting, but only when supported by strong pretrained models.

Surprisingly, we find that these observations hold even when the number of time steps is increased to 200, a large increase compared to current benchmarks, while normalizing the effective total computation accordingly. This suggests that existing CL literature is particularly suited for settings where memory is limited, and less practical in scenarios having limited computational budgets.

\section{Continual Learning with Limited Compute}
\vspace{-0.1cm}
\label{sec:budgted_cl}

\subsection{Problem Formulation}
\vspace{-0.1cm}
We start by first defining our proposed setting of \textit{computationally budgeted} continual learning. Let $\mathcal{S}$ be a stream revealing data sequentially over time steps. At each time step $t \in \{1,2,\dots,\infty\}$, the stream $\mathcal{S}$ reveals $n_t$ image-label pairs $\{(x_i^t,y_i^t)\}_{i=1}^{n_t} \sim \mathcal{D}_j$ from distribution $\mathcal{D}_j$ where $j \in \{1,\dots,t\}$. In this setting, we seek to learn a function $f_{\theta_t} : \mathcal{X} \rightarrow \mathcal{Y}_t$ parameterized by $\theta_t$ that maps images $x\in \mathcal{X}$ to class labels $y \in \mathcal{Y}_t$, where $\mathcal{Y}_t = \bigcup^t_{i=1}\mathcal{Y}_i$, which aims to correctly classify samples from any of the previous distributions $\mathcal{D}_{j \leq t}$. In general, there are no constraints on the incoming distribution $\mathcal{D}_j$, \eg, the distribution might change after every time step or it may stay unchanged for all time steps. The size of the revealed stream data $n_t$ can generally change per step, \eg, the rate at which users upload data to a server. The unique aspect about our setting is that at \emph{every time step $t$, a computational budget $\mathcal{C}_t$} is available for the CL method to update the parameters from $\theta_{t-1}$ to $\theta_t$ in light of the new revealed data. Due to the inexpensive costs associated with memory storage, in our setting, we assume that CL methods in our setting can have full access to all previous samples $\mathcal{T}_t = \cup_{r=1}^{t} \{(x_i^r,y_i^r)\}_{i=1}^{n_r}$.\footnote{We discuss in the Appendix the privacy arguments often used towards restricting the memory.} 
However, as will be discussed later, while all samples can be stored, they cannot all be used for training due to the constrained computation imposing an implicit memory restriction. 

\begin{table*}[]
\scriptsize
    \centering
    \resizebox{0.775\textwidth}{!}{
    \begin{tabular}{llcccccc} \toprule
    Dir. & Reference & Applicability & \multicolumn{5}{c}{Components} \\
     &  & (our setup) & Distillation & MemUpdate & MemRetrieve & FC Correction & Others\\\midrule
     & Naive & $\checkmark$ & - & Random & Random & - & - \\ \hline
  \multirow{6}{*}{\rotatebox[origin=c]{90}{Distillation}}  & iCARL~\cite{rebuffi2017icarl}& $\checkmark$ & \textbf{BCE} & \textbf{Herding} & Random & - & \textbf{NCM} \\
     & LUCIR~\cite{hou2019learning} & $\checkmark$  & \textbf{Cosine} & Herding & MargRank & \textbf{CosFC} & NCM \\
     & PODNet~\cite{douillard2020podnet} & $\checkmark$  & \textbf{POD} & Herding & Random & \textbf{LSC} & Imprint,NCM\\
     & DER~\cite{buzzega2020dark} & $\checkmark$  & \textbf{MSE} & Reservoir & Random & - & - \\
      
     &  C$\mbox{O}^2$L \cite{cha2021co2l} & $\times$ & \textbf{IRD} & Random & Random & \textbf{Asym.SupCon} & - \\
     &  SCR \cite{mai2021supervised} & $\checkmark$ & - & Reservoir & Random & \textbf{SupCon} & NCM \\ \hline
   \multirow{11}{*}{\rotatebox[origin=c]{90}{Sampling}}  & TinyER~\cite{chaudhry2019continual}& $\checkmark$ & -  & \textbf{\begin{tabular}{c}FIFO,KMeans,Reservoir\end{tabular}} & - &  - & -\\
     & GSS~\cite{aljundi2019gradient} & $\times$ & - & \textbf{GSS} & Random & - & -\\
     &   MIR~\cite{aljundi2019online} & $\times$ & - & Reservoir & \textbf{MIR} & - & - \\ 
     & GDumb~\cite{prabhu2020gdumb}& $\checkmark$ & - & \textbf{Balanced} & Random & - & \textbf{MemOnly} \\
     & Mnemonics~\cite{liu2020mnemonics} & $\times$ & - & \textbf{Mnemonics} & - & - & BalFineTune  \\ 
     & OCS\cite{yoon2021online}& $\times$ & - & \textbf{OCS} & Random & - & - \\
     & InfoRS~\cite{sun2022information} & $\times$ & MSE & \textbf{InfoRS} & Random & - & - \\
     & RMM\cite{liu2021rmm}  & $\times$ & - & \textbf{RMM} & - & - & - \\
    & ASER \cite{shim2021online} & $\times$ & - & \textbf{SV} & \textbf{ASV} & - & - \\
    & RM \cite{bang2021rainbow} & $\checkmark$ & - & \textbf{Uncertainty} & Random & - & AutoDA  \\
     & CLIB~\cite{koh2021online}& $\times$ & - & \textbf{Max Loss} & Random & - & \begin{tabular}{c} MemOnly,AdaLR\end{tabular}  \\\hline
   \multirow{5}{*}{\rotatebox[origin=c]{90}{FC Layer}} & BiC~\cite{wu2019large} & $\times$ & CrossEnt & Random & Random & \textbf{BiC} & - \\
     & WA~\cite{zhao2020maintaining} & $\times$ & CrossEnt & Random & Random &\textbf{WA} & - \\
    & SS-IL \cite{ahn2021ss} & $\times$ & TKD & Random & Balanced & \textbf{SS} &  -\\
     & CoPE\cite{de2021continual}& $\checkmark$ & - & Balanced & Random & \textbf{PPPLoss} & - \\
     & ACE~\cite{caccia2022new} & $\checkmark$ & - & Reservoir & Random & \textbf{ACE} & - \\ \bottomrule
    \end{tabular}}
    \vspace{-0.15cm}
    \caption{\textbf{Primary Directions of Progress in CL.} Analysis of recent replay-based systems, with \textbf{bold} highlighting the primary contribution. We observe that there are three primary directions of improvement. ``App." denotes the applicability to our setting based on whether they are scalable to large datasets and applicable beyond the class-incremental stream.}
    \vspace{-0.60cm}
    \label{tab:approaches}
\end{table*}

\subsection{Key Differences with Prior Art}
\vspace{-0.1cm}

(\textbf{1}) \textit{Tasks}: In most prior work, CL is simplified to the problem of learning a set of non-overlapping tasks, \ie, distributions, \emph{with known boundaries} between them \cite{rebuffi2017icarl,lopez2017gradient,chaudhry2018riemannian}. 
In particular, the data of a given distribution $\mathcal{D}_j$ is given all at once for the model to train. This is as opposed to our setup, where there is no knowledge about the distribution boundaries, since they are often gradually changing and not known a priori. As such, continual learning methods cannot train only just before the distribution changes.

(\textbf{2}) \textit{Computational Budget}: A key feature of our work is that, per time step, CL methods are given a fixed computational budget $\mathcal{C}_t$ to train on $\{(x_i^t,y_i^t)\}_{i=1}^{n_t}$. For ease, we assume throughout that $\mathcal{C}_t = \mathcal{C} ~\forall t$, and that $n_t = n \forall t$. Although $\mathcal{C}$ can be represented in terms of wall clock training time, for a given $f_\theta$ and stream $\mathcal{S}$, and comparability between GPUs, we state $\mathcal{C}$ in terms of the number of training iterations instead. This avoids hardware dependency or suboptimal implementations when comparing methods. This is unlike prior work, which do not put hard constraints on compute per step \cite{rebuffi2017icarl,hou2019learning,buzzega2020dark} giving rise to degenerate but well-performing algorithms such as GDumb \cite{prabhu2020gdumb}. Concurrent works \cite{ghunaim2023real, bornschein2022nevis} restrict the computational budget, however, they operate in a setup with constrained memory which significantly affects performance of CL methods.

(\textbf{3}) \textit{Memory Constraints}: Prior work focuses on a fixed, small memory buffer for learning and thereof proposing various memory update strategies to select samples from the stream. We assume that all the samples seen so far can be stored at little cost. However, given the restricted imposed computation $\mathcal{C}$, CL methods cannot revisit or learn from all stored samples. For example, as shown in Figure \ref{fig:effective_epochs}, consider performing continual learning on ImageNet2K, composed of 1.2M samples from ImageNet1K and 1.2M samples from ImageNet21K forming 2K classes, which will be detailed later, over $20$ time steps, where the stream reveals sequentially $n = 60$K images per step. Then, under a computation budget of $8000$ iterations, the model cannot revisit more than $50\%$ of all seen data at any given time step, \ie 600K samples. Our proposed setting is closer to realistic scenarios that cope with high-throughput streams, similar to concurrent work \cite{prabhu2023online}, where \textit{computational bottlenecks impose implicit constraints on learning from past samples that can be too many to be revisited during training}.

\subsection{Constructing the Stream} \vspace{-0.1cm}
\noindent We explore three stream settings in our proposed benchmark, which we now describe in detail.

(\textbf{1}) \textit{Data Incremental Stream}: In this setting, there is no restriction on the incoming distribution $\mathcal{D}_j$ over time that has not been well-explored in prior works.
We randomly shuffle all data and then reveal it sequentially over steps, which could lead to a varying distribution $\mathcal{D}_j$ over steps in which there are no clear distribution boundaries.

(\textbf{2}) \textit{Time Incremental Stream}: In this setting, the stream data is ordered by the upload timestamp to a server, reflecting a natural distribution change $\mathcal{D}_j$ across the stream as it would in real scenarios. There is a recent shift toward studying this ordering as apparent in recent CL benchmarks, \eg, CLEAR \cite{lin2021clear}, Yearbook \cite{yao2022wild} and FMoW \cite{yao2022wild}, NEVIS22 \cite{bornschein2022nevis}, Continual YFCC100M \cite{cai2021online} and Continual Google Landmarks V2 \cite{prabhu2023online}.

(\textbf{3}) \textit{Class Incremental Stream}: For completeness, we consider this classical setting in the CL literature. Each of the distributions $\mathcal{D}_j$ represents images belonging to a set of classes different from the classes of images in any other distribution $\mathcal{D}_{i \neq j}$. We benchmark these three settings using a large-scale dataset that will be detailed in the Experiments.

\section{Dissecting Continual Learning Systems}
\vspace{-0.1cm}

Continual learning methods typically propose a system of multiple components that jointly help improve learning performance. For example, LUCIR\cite{hou2019learning} is composed of a cosine linear layer, a cosine distillation loss function, and a hard-negative mining memory-based selector. In this section, we analyze continual learning systems and dissect them into their underlying components. This helps to analyze and isolate the role of different components under our budgeted computation setting and helps us to understand the most relevant components.  
In Table \ref{tab:approaches}, we present the breakdown of novel contributions that have been the focus of recent progress in CL. 
The columns indicate the major directions of change in the CL literature. Overall, there have been three major components on which advances have focused, namely distillation, sampling, and FC layer correction. These three components are considered additions to a naive baseline that simply performs uniform sampling from memory. We refer to this baseline as Naive in Table \ref{tab:approaches}.

(\textbf{1}) \textit{Distillation}: One popular approach towards preserving model performance on previous distributions has been through distillation. It enables student models, \ie, current time step model, to learn from a teacher model, \ie, one that has been training for many time steps, through the logits providing a rich signal. 
In this paper, we consider four widely adopted distillation losses, namely, Binary CrossEntropy (BCE) \cite{rebuffi2017icarl}, CrossEntropy \cite{wu2019large,zhao2020maintaining,liu2020mnemonics}, Cosine Similarity (Cosine)\cite{hou2019learning}, and Mean Square Error (MSE)\cite{buzzega2020dark,sun2022information} Loss.

(\textbf{2}) \textit{Sampling}: Rehearsing samples from previous distributions is another popular approach in CL. However, sampling strategies have been used for two objectives. Particularly when access to previous samples is restricted to a small memory, they are used to select which samples from the stream will update the memory (MemUpdate) or to decide on which memory samples are retrieved for rehearsal (MemRetrieve). 
In our unconstrained memory setup, simply sampling uniformly over the joint data of past and current time step data (as in Naive) exposes a particular shortcoming. When training for a large number of time steps, uniform sampling reduces the probability of selecting samples from the current time step. For that, we consider various sampling strategies, \eg, recency sampling \cite{lin2021clear} that biases toward sampling current time step data, and FIFO \cite{chaudhry2019continual, cai2021online} that exclusively samples from the current step. We do not consider Reservoir, since it approximates uniform sampling in our setup with no memory restrictions. In addition to benchmarking the sampling strategies mentioned above, we also consider approaches that evaluate the contribution of each memory sample to learning \cite{toneva2018empirical}. For example, Herding \cite{rebuffi2017icarl}, K-Means \cite{chaudhry2019continual}, OCS \cite{yoon2021online}, InfoRS \cite{sun2022information}, RM  \cite{bang2021rainbow}, and GSS \cite{aljundi2019gradient} aim to maximize diversity among samples selected for training with different metrics. MIR \cite{aljundi2019online}, ASER\cite{shim2021online}, and CLIB \cite{koh2021online} rank the samples according to their informativeness and select the top-$k$. Lastly, balanced sampling \cite{prabhu2020gdumb,de2021continual,chrysakis2020online} select samples such that an equal distribution of classes is selected for training. In our experiments, we only consider previous sampling strategies that are applicable to our setup and compare them against Naive.

(\textbf{3}) \textit{FC Layer Correction}: It has been hypothesized that the large difference in the magnitudes of the weights associated with different classes in the last fully connected (FC) layer is among the key reasons behind catastrophic forgetting \cite{wu2019large}. There has been a family of different methods addressing this problem. 

These include methods that improve the design of FC layers, such as CosFC \cite{hou2019learning}, LSC\cite{douillard2020podnet}, and PPP \cite{de2021continual}, by making the predictions independent of their magnitude. 
Other approaches such as SS-IL \cite{ahn2021ss} and ACE \cite{zeno2018task, mai2022online, caccia2022new} mask out unaffected classes to reduce their interference during training. In addition, calibrating the FC layer in post-training, \eg, 
BiC \cite{wu2019large}, WA \cite{zhao2020maintaining}, and IL2M \cite{belouadah2019il2m} is widely used. Note that the calibration techniques are only applicable to the class-incremental setup. We benchmark existing methods applicable to our setting against the Naive approach that does not implement any FC layer correction.

(\textbf{4}) \textit{Model Expansion Methods}:
Several works attempt to adapt the model architecture according to the data. This is done by only training part of the model \cite{mallya2018packnet,mallya2018piggyback,abati2020conditional,aljundi2017expert, rajasegaran2019random} or by directly expanding the model when data is presented \cite{rusu2016progressive,yan2021dynamically,yoon2017lifelong,zhang2020side,wang2017growing,rebuffi2017learning}. 
However, most of the previous techniques in this area do not apply to our setup. Most of this line of work \cite{mallya2018packnet,mallya2018piggyback,rebuffi2017learning} assumes a task-incremental setting, where at every time step, new samples are known to what set of classes they belong, \ie, the distribution boundaries are known, even at test time. To overcome these limitations, newer methods \cite{rajasegaran2020itaml,abati2020conditional} use a bilevel prediction structure, predicting the task at one level and the label within the task at the second level. They are restricted to the class-incremental setting as they assume each task corresponds to a set of non-overlapping classes. We seek to understand the limitation of partial retraining in a network; hence, instead, we compare Naive against a setting where only the FC layer is being trained, thus minimally training the network per time step. In addition, we examine the role of pretraining which has recently become a widely popular direction for exploration in continual learning \cite{wu2022class}.
\section{Experiments}
\vspace{-0.1cm}
\label{experiments}

We first start by detailing the experimental setup, datasets, computational budget $\mathcal{C}$, and evaluation metrics for our large-scale benchmark. We then present the main results evaluating various CL components, followed by extensive analysis.

\subsection{Experimental Setup and Details}
\vspace{-0.1cm}
\label{sec:setup}

\textbf{Model.} We use a standard ResNet50 following prior work on continual learning \cite{cai2021online}. The model is ImageNet1K pre-trained used as a backbone throughout all experiments.

\textbf{Datasets.} We conduct experiments using two large-scale datasets, namely  ImageNet2K and Continual Google Landmarks V2 (CGLM). We construct ImageNet2K by augmenting ImageNet1K with $1.2$M images from ImageNet21K \cite{deng2009imagenet}, thus, adding $1$K new non-overlapping classes with ImageNet1K amounting to a total of $2$K classes. 

(\textbf{1}) \textit{Data Incremental ImageNet2K}: The stream is constructed by randomly shuffling the set of images from the $1$K classes of ImageNet21K, by doing so, there is no knowledge of the distribution boundaries. The model continually learns on this set of images, while ImageNet1K is available in memory. CL methods are expected to learn both the new classes from the stream while maintaining the performance on ImageNet1K. We refer to this setting as \emph{DI-ImageNet2K}.

(\textbf{2})  \textit{Class Incremental ImageNet2K}: Similar to the above defined DI-ImageNet2K, ImageNet1K is available in memory and the $1$K classes of ImageNet21K are presented sequentially by the stream but in a class incremental setting. We refer to this setting as \emph{CI-ImageNet2K}.  

(\textbf{3}) \textit{Time Incremental Google Landmarks V2 (CGLM)}: In this setting, the stream consists of data from the CGLM dataset ordered according to the timestamps of the images mimicking a natural distribution shift. Note that ImageNet1K is not considered as part of the evaluation. We refer to this setting simply as \emph{CGLM}.

Throughout, unless stated otherwise, the stream reveals data incrementally over 20 time steps. This amounts to a per step stream size of $n = 60$K for the CI-ImageNet2K and DI-ImageNet2K settings, and $n = 29$K for the CGLM setting. More details on the construction of datasets is given in the Appendix along with the sample orders.

\begin{figure}[t]
    \centering
    \hspace{-0.4cm}\includegraphics[width=0.45\textwidth]{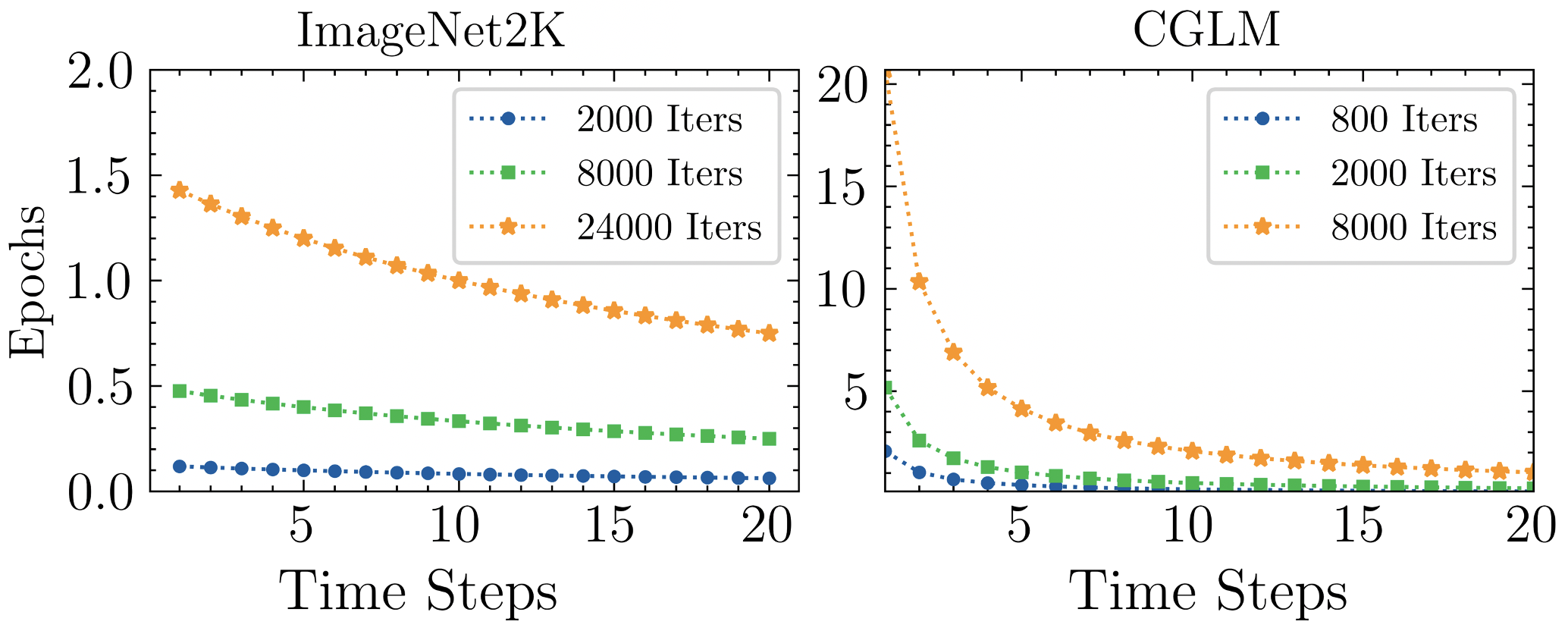}
    \vspace{-0.2cm}
    \caption{\textbf{Effective Training Epochs Per Time Step.} Our default setting sets a total training budget over all $20$ time steps of $8000$ and $2000$ iterations for ImageNet2K and CGLM ,respectively, with a per iteration batch size of $\mathcal{B} = 1500$. Effectively, this reflects to training on 25-50\% of the stored data, except in the first few time steps on CGLM. Note that for ImageNet2K, we assume that ImageNet1K of $1.2$M samples is available in memory.}
    \vspace{-0.60cm}
    \label{fig:effective_epochs}
\end{figure}

\begin{figure*}[htp!]
    \centering
    \includegraphics[width=\textwidth]{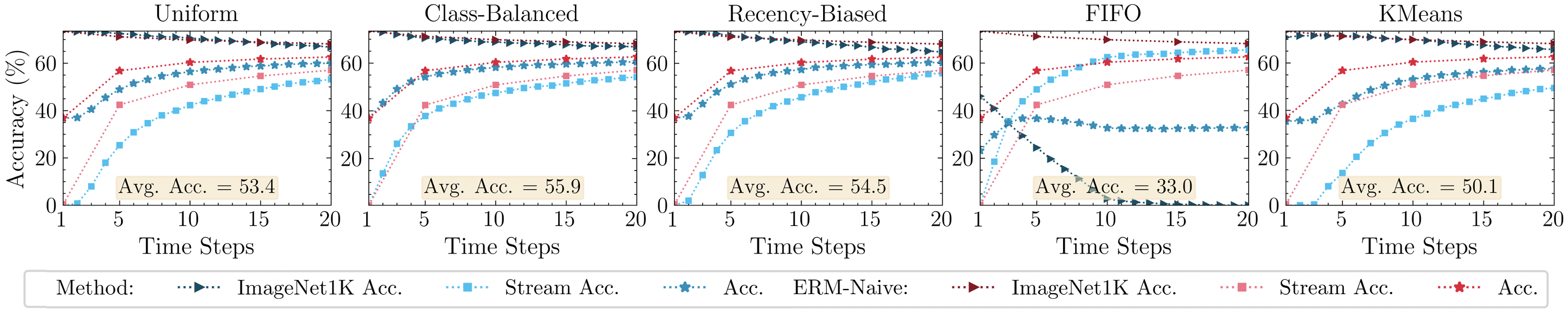}
    \vspace{-0.7cm}
    \caption{\textbf{DI-ImageNet2K ($400$ Iterations). } ERM-Naive, a non-continual learning algorithm, is compared against inexpensive sampling strategies (first four plots) with $400$ training iterations and the costly KMeans (fifth plot) with $200$ iterations. All CL methods perform similarly but worse than ERM-Naive. This is the case for FIFO that suffers from forgetting the KMeans due to its expensive nature. ImageNet2K experiments performance can be decomposed into (i) accuracy on classes seen during pre-training on ImageNet1K and (ii) accuracy on newly seen classes in ImageNet2K, allowing analysis of forgetting old classes and learning newly introduced classes.}
    \label{fig:datainc_sampling}
    \vspace{-0.35cm}
\end{figure*}

\begin{figure*}[htp!]
    \centering
    \includegraphics[width=\textwidth]{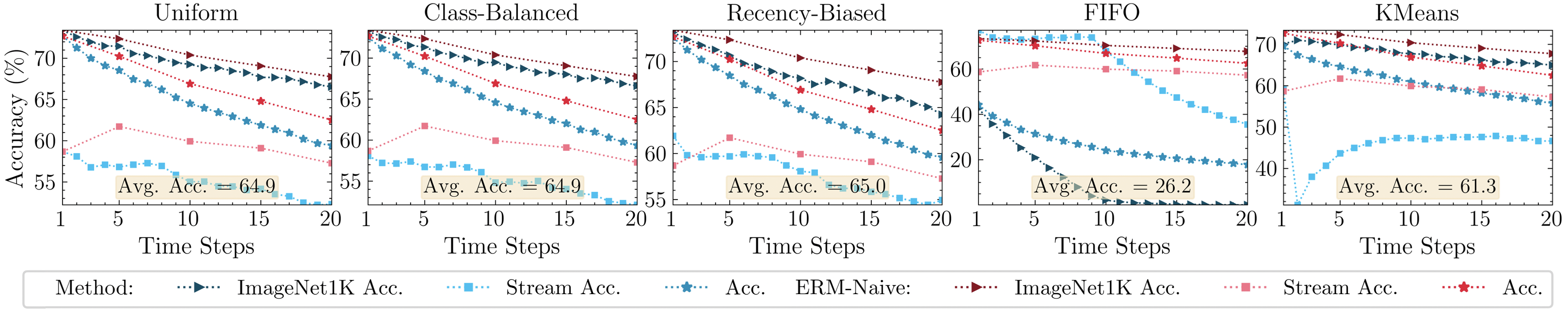}
\vspace{-0.7cm}
    \caption{\textbf{CI-ImageNet2K ($400$ Iterations). } Similarly, ERM-Naive is compared on CI-ImageNet2K. Both FIFO, which suffers from forgetting, and KMeans due to its expensive nature, struggle to compete against simpler inexpensive methods as Class-Balanced. However, overall, all other methods perform very similarly with no clear advantage. ImageNet2K experiments performance can be decomposed into (i) accuracy on classes seen during pre-training on ImageNet1K and (ii) accuracy on newly seen classes in ImageNet2K, allowing analysis of forgetting old classes and learning newly introduced classes.}
    \label{fig:clsinc_sampling}
    \vspace{-0.35cm}
\end{figure*}
\begin{figure*}[htp!]
    \centering
   
    \includegraphics[width=\textwidth]{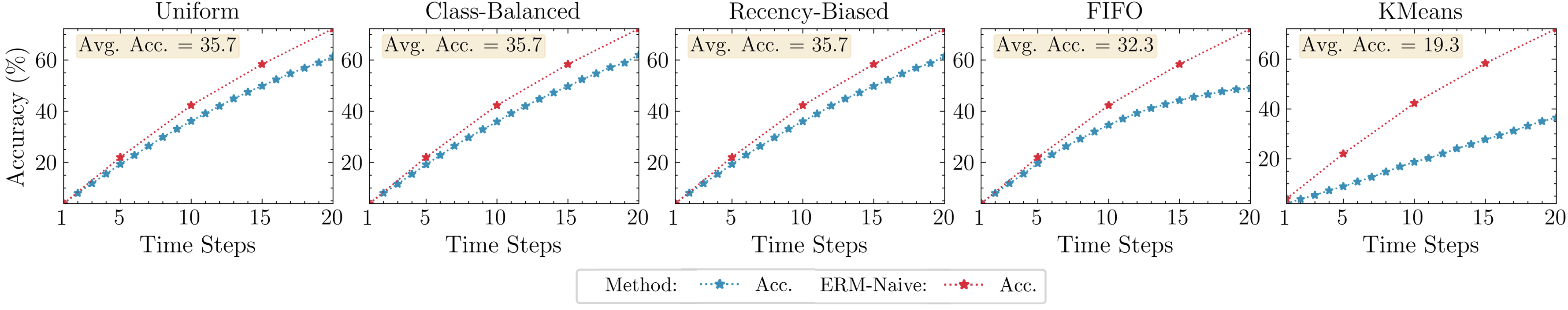}
    \vspace{-0.7cm}
\caption{\textbf{CGLM ($100$ Iterations). } All inexpensive methods perform overall similarly with the exception for KMeans due to its expensive nature. This highlights that simplicity is key under a budgeted continual learning setting. CGLM is not an extension of ImageNet1K and involves a different task: landmark classification. Hence, we measure only the stream accuracy resulting in two lines instead of six.}
\vspace*{-0.50cm}
    \label{fig:nds_sampling}
\end{figure*}

\begin{figure}[htp!]
    \centering
    \includegraphics[width=0.9\linewidth]{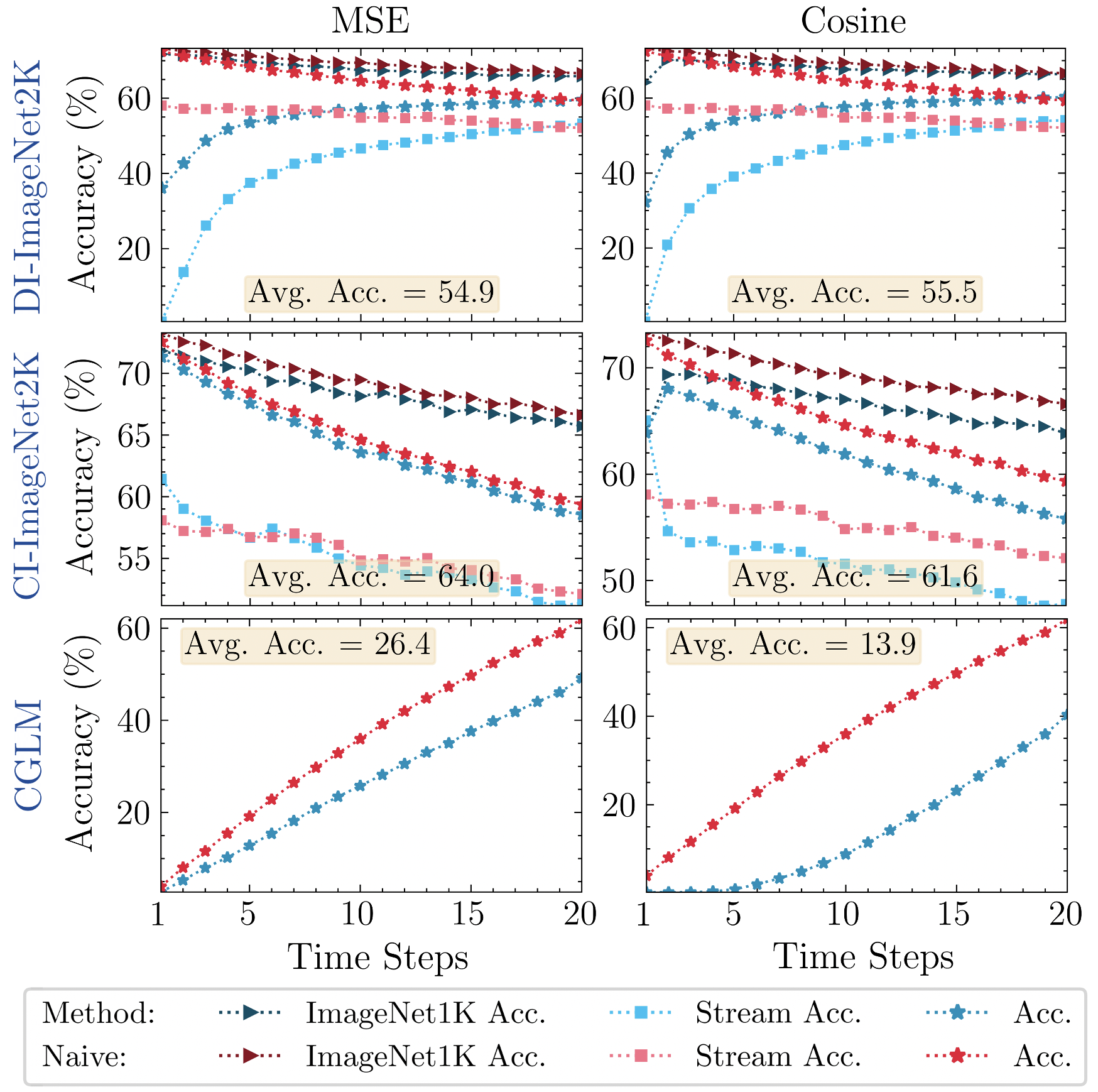}
    \vspace{-0.2cm}
    \caption{ \textbf{Distillation in Data and Class Incremental Settings.} Naive, which does not employ any distillation loss, outperforms all distillation methods (MSE, and Cosine) across all three settings.}
    \vspace{-0.60cm}
    \label{fig:distillation_datainc}
\end{figure}

\textbf{Computational Budget.} We set the computational budget $\mathcal{C}$ to 400 training iterations per time step ($8000 = 20 \text{ time steps} \times 400$) for ImageNet2K, \ie, DI-ImageNet2K and CI-ImageNet2K, and set $\mathcal{C}$ to 100 training iterations for CGLM. In each iteration, a batch of images is used to update the model in training where we set the training batch size $\mathcal{B}$ to $1500$. The choice of $\mathcal{C}$ is made such that it corresponds to training on at most $25-50\%$ of all observed data at any given step. For example, as highlighted in Figure \ref{fig:effective_epochs} for ImageNet2K, time step $t = 5$. corresponds to training only on about $40\%$ of the complete observed data at this step, \ie, $\nicefrac{400 \times 1500}{1.2M  + 5 \times 60K} \approx 0.4$ of an epoch where $1.2$M denotes the ImageNet1K samples. Furthermore, we set $\mathcal{C}$ to 100 iterations for CGLM, since the dataset contains \nicefrac{1}{4} of the total data in ImageNet2K.  Note that after 20 time steps on CGLM, the data that would have been seen is $20 \times 29K$ images, as opposed to $1.2M  + 20 \times 60K$ images for ImageNet2K experiments.

\textbf{Metrics.} We report the accuracy (Acc) on a separate test set after training at each time step. This test set simply comprises the joint test set for all classes seen up to the current time step. Moreover, for ImageNet2K, we decompose the test accuracy into the accuracy on ImageNet1K (ImageNet1K Acc), which measures forgetting, and the accuracy on the stream (Stream Acc), which measures adaptation. For GCLM, we only report stream accuracy.

\noindent\textbf{Training Details.} We use SGD as an optimizer with a linear learning rate schedule and a weight decay of $0$. We follow standard augmentation techniques. 
All experiments were run on the same A100 GPU. For a fair comparison, we fix the order of the samples revealed by the stream $\mathcal{S}$ in all experiments on a given dataset and comparisons.

\noindent We summarize all the settings with all the benchmark parameters in the first part of Table \ref{tab:setup_deets}.

\subsection{Budgeted Continual Learning}
\vspace{-0.15cm}
In this section, we investigate the effectiveness of the three main directions studied in the CL literature, namely sampling strategies, distillation, and FC layer correction.

\noindent \textbf{1. Do Sampling Strategies Matter?} We evaluate seven sampling strategies that govern the construction of the training batch from memory. These strategies are grouped into two categories based on their computational cost. Inexpensive sampling methods include Uniform, Class-Balanced, Recency-Biased and FIFO sampling. On the other hand, costly sampling strategies include KMeans, Max Loss, and Uncertainty loss sampling. 

To normalize for the effective $\mathcal{C}$ due to the overhead of associated extra forward passes to decide on the sampling, costly sampling strategies are allowed $\nicefrac{\mathcal{C}}{2}$ training iterations, where the exact calculation is left for the Appendix. That is to say, costly sampling strategies perform $200$ training iterations for ImageNet2K and $50$ training iterations for CGLM as the rest of the budget is for the extra forward passes. We report the performance of the five sampling strategies consisting of the inexpensive and the best performing costly sampling strategy (KMeans),  presented in shades of blue, in Figures \ref{fig:datainc_sampling}, \ref{fig:clsinc_sampling}, and \ref{fig:nds_sampling} for DI-ImageNet2K, CI-ImageNet2K, and CGLM, respectively. Other methods are listed in the Appendix due to lack of space. We compare against a non-continual learning oracle that performs classical empirical risk minimization at every step on $\mathcal{T}_t = \cup_{r=1}^t \{(x_i^r,y_i^r)\}_{i=1}^{n_r}$ with a computational budget of $\mathcal{C} \times t$, which we refer to as ERM-Naive; this is as opposed to the previously mentioned continual learning methods that have only $\mathcal{C}$ per step $t$ spent equally over all steps. ERM-Naive acts as a training method with hindsight, spending the complete computational budget at once after collecting the full dataset. This acts as a very strong baseline against all continual learning methods. We report it in shades of red in the same figures. We also report the average accuracy, averaged over all time steps, for each sampling method in the yellow box in each figure.

\begin{table}[t]
    \centering
    \scriptsize
    \begin{tabular}{l|cc}\toprule
        \textbf{Attributes} & \textbf{ImageNet2K} & \textbf{CGLM} \\ \hline
        Initial memory & ImageNet1K & \{\} \\
        Initial memory size & 1.2M & 0 \\
        Per step stream size $n$ & 60K & 29K \\
        Time steps & 20 & 20 \\
        Stream size & 1.2M & 58K \\
        Size of data by the last time step & 2.4M & 58K \\
        Stream & \begin{tabular}{c} Class incremental \\ Data incremental \end{tabular} & Time incremental\\
        $\#$ iterations per time step $\mathcal{C}$ & 400 & 100 \\
        Training batch size $\mathcal{B}$ & 1500 & 1500 \\
        Metrics & \begin{tabular}{c} Acc on ImageNet1K \\Acc on Stream \end{tabular} & Acc on Stream\\
        \hline
        Eq. Distillation Iters & 267 & 67 \\
        Eq. Sampling Iters & 200 & 100 \\
        Eq. FC Correction Iters & 400 & 100 \\ \hline
        Iters per $t$ (Sensitivity) & 100, 1200 & 40, 400\\
        Time Steps (Sensitivity) & 50, 200 & 50, 200\\\bottomrule
    \end{tabular}
    \vspace{-0.2cm}
    \caption{\textbf{Experimental Details.} The first block shows the various considered settings in the experiments section. The second block denotes the effective training iterations $\mathcal{C}$ for each class of methods due to their over head extra computation. The last block details the setup for our sensitivity analysis.}
    \vspace{-0.60cm}
    \label{tab:setup_deets}
\end{table}

\noindent\textbf{Conclusion.} First, we observe that the top inexpensive sampling strategies perform very similarly to each other. This is consistent across settings, CI-ImageNet2K, and CGLM, on both ImageNet1K accuracy and Stream Accuracy. There are some advantages for Class-Balanced over other sampling strategies, \eg, gaining an average accuracy of 2.5\% over Uniform in DI-ImageNet2K. However, sampling strategies such as FIFO completely forget ImageNet1K (dark blue line), leading to poor performance over all three settings. Interestingly, costly sampling strategies perform significantly worse in CL performance over the simple Uniform sampling when subjected to an effectively similar computational budget. This observation is different from previous settings \cite{masana2020class}, as the additional computational overhead of costly sampling does not seem worthwhile to improve performance.

\begin{figure*}[htp!]
    \centering
    \vspace*{-0.1cm}
    \includegraphics[width=0.75\textwidth]{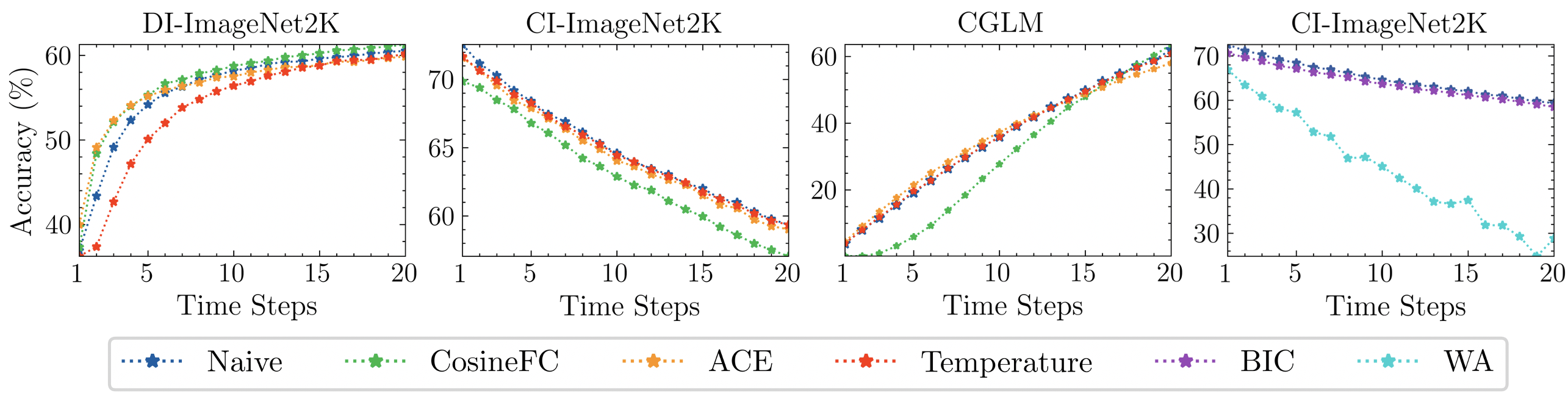}
    \vspace{-0.2cm}
    \caption{\textbf{FC Layer Correction.} Even though loss functions (CosineFC and ACE) might outperform Naive in the first few time steps, eventually Naive catches up. Overall, Naive consistently outperforms all considered calibration methods too, namely, BIC and WA.}
    \vspace{-0.35cm}
    \label{fig:lastlayer}
\end{figure*}

\noindent\textbf{2. Does Distillation Matter?} We evaluate four well-known distillation losses in our benchmark, namely, Cosine, CrossEntropy, BCE, and MSE losses. Given that Class-Balanced is a simple inexpensive sampling procedure that performed slightly favorably, as highlighted in the previous section, we use it as a default sampling strategy from now onward, where the number of samples used per training step is equal over all classes. We refer to this basic approach with a cross entropy loss as Naive. To fairly factor in the overhead of an additional forward pass, distillation approaches are allowed $\nicefrac{2\mathcal{C}}{3}$ iterations compared to Naive with $\mathcal{C}$ training iterations. That is, the distillation losses perform $267$ iterations for ImageNet2k and $67$ iterations for CGLM compared to $400$ and $100$ iterations for Naive. We report the results for Cosine and MSE on DI-ImageNet2K, CI-ImageNet2K, and CGLM datasets in the first, second, and third rows of Figure \ref{fig:distillation_datainc}, respectively. Other methods are left for the Appendix due to lack of space. Distillation methods are shown in shades of blue, whereas Naive is shown in shades of red. We report the average accuracy, averaged over all time steps, for each distillation method in the yellow box in each figure.

\noindent\textbf{Conclusion.} In all three settings, distillation methods underperform compared to Naive. Even in ImageNet1K Acc, which measures forgetting, Naive performs similarly or slightly better than all distillation methods in DI-ImageNet2K and CI-ImageNet2K streams. The results in Figure \ref{fig:distillation_datainc} show that the top distillation methods, such as MSE, perform only slightly worse compared to Naive ($54.9$ vs $55.9$ on DI-ImageNet2K and $64$ vs $64.9$ on CI-ImageNet2K). However, in CGLM they perform significantly worse ($26.4$ compared to $35.7$) due to the limited iterations. We attribute this to the fact that distillation methods often require a larger number of training samples, and thereof a large enough computational budget per time step.

\begin{figure}[t]
    \centering
    \vspace{-0.15cm}
    \includegraphics[width=0.4\textwidth]{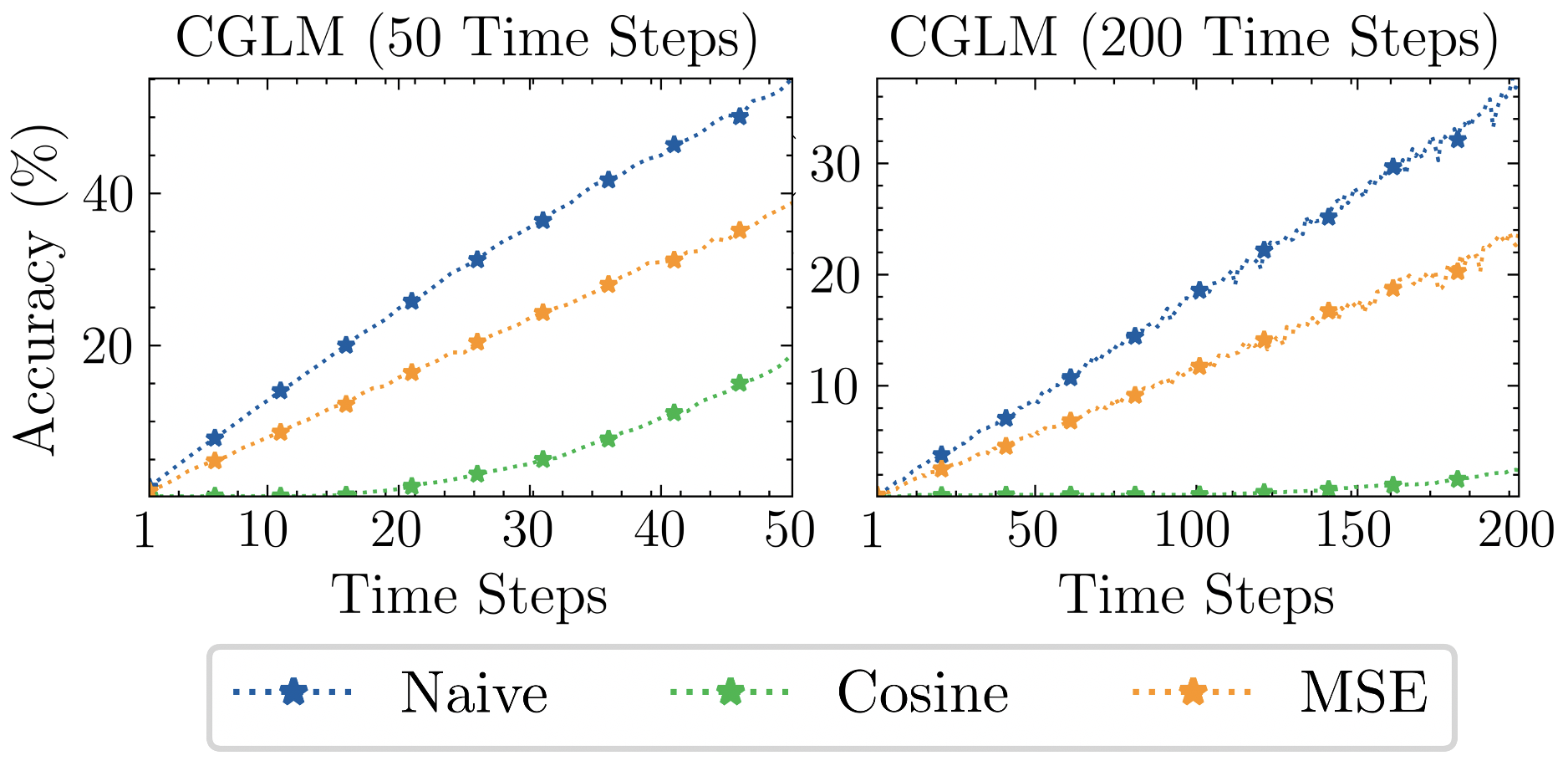}
    \vspace{-0.25cm}
    \caption{\textbf{CGLM Distillation with Different Number of Time Steps.} Under larger number of time steps, where total number of iterations is normalized accordingly, Naive outperforms distillation in both settings, namely, $50$ and $200$ time steps.}
    \label{fig:difftimestepscglm}
    \vspace{-0.4cm}
\end{figure}

\noindent \textbf{3. Does FC Layer Correction Matter?} We evaluate five FC layer correction approaches from two different families. A family of methods that modifies the FC layer directly, including CosineFC \cite{hou2019learning} and ACE\footnote{We treat all seen samples as incoming samples to test this case, diverging from the original ACE Loss. The ACE Loss collapses to Crossentropy, as new samples form a tiny fraction of all past seen data.} \cite{zeno2018task,mai2022online,caccia2022new}. The other family of methods applies post-training calibration including BiC \cite{wu2019large}, WA \cite{zhao2020maintaining}, along with temperature scaling \cite{guo2017calibration}. All methods employ Class-Balanced as a sampling strategy and compare against Naive (Class-balanced with cross entropy loss) with no corrections in FC layer. The first three subplots of Figure \ref{fig:lastlayer} correspond to comparisons of direct FC layer modification methods against Naive on DI-ImageNet2K, CI-ImageNet2K, and CGLM. Since calibration methods tailored for Class Incremental settings, in the rightmost plot of Figure \ref{fig:lastlayer}, we report comparisons with Naive on CI-ImageNet2K. Since all FC layer corrections are with virtually no extra cost, the number of training iterations per time step is set to $\mathcal{C}$, \ie, $400$ for ImageNet2K and $100$ for CGLM.
 
\noindent\textbf{Conclusion.} No method consistently outperforms Naive in computationally budgeted continual learning. The first family of methods helps in DI-ImageNet2K, particularly in the initial steps due to class imbalance, but no method outperforms Naive in the CI-ImageNet2K set-up. Calibration-based methods, such as BIC, are somewhat competitive with Naive, but WA fails. Surprisingly, even under various FC correction approaches, all methods fail to outperform Naive in computationally budgeted continual learning.

\begin{figure}[t]
    \centering
    \includegraphics[width=0.4\textwidth]{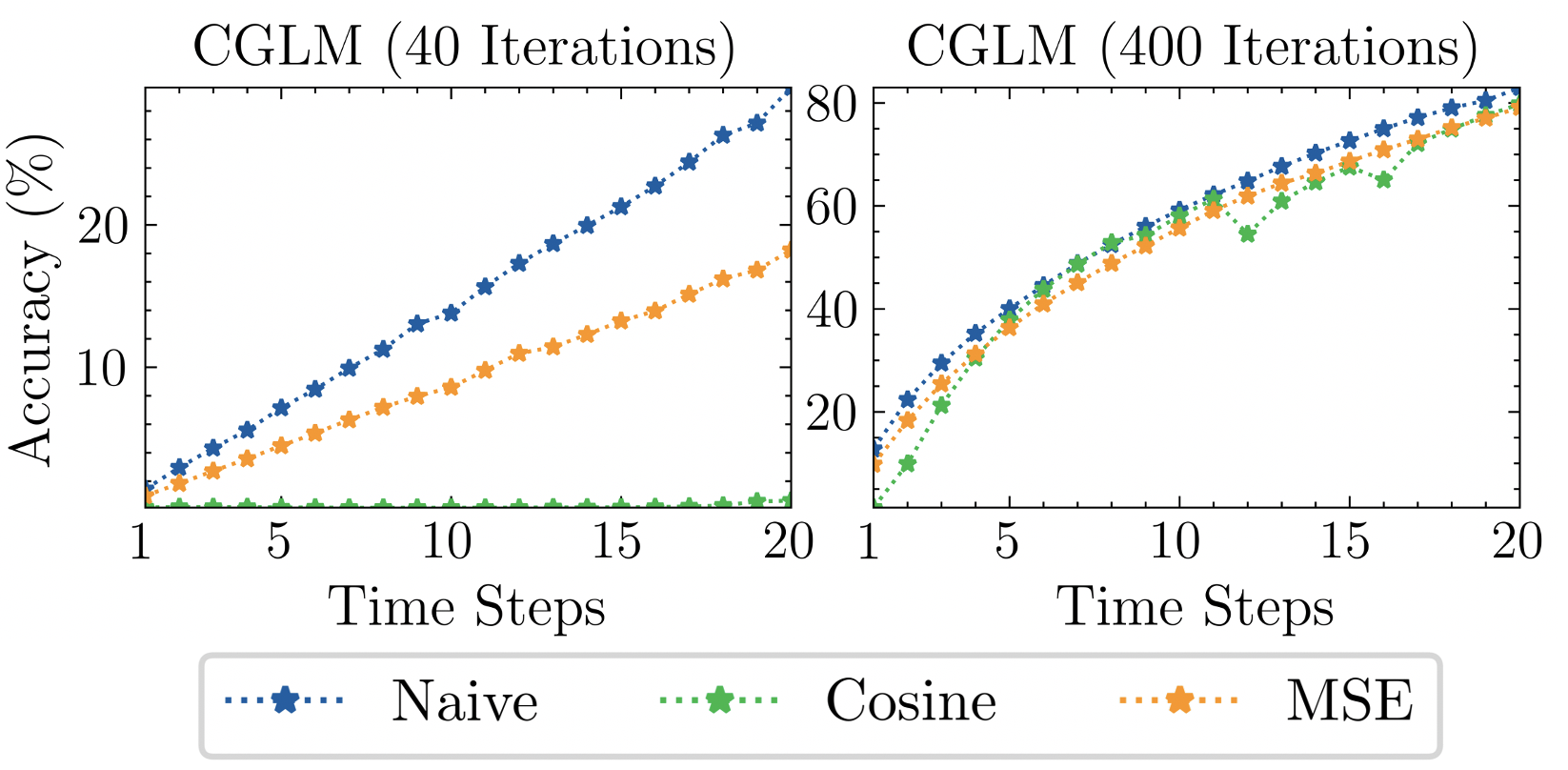}
    \vspace{-0.25cm}
    \caption{\textbf{CGLM Distillation with Different Computational Budgets.} Naive outperforms distillation methods under the restricted $40$ and the larger $400$ iterations (originally $100$). Distillation methods become competitive when enough budget is available.}
    \label{fig:diffstepscglm}
    \vspace{-0.4cm}
\end{figure}
\vspace{-0.15cm}
\subsection{Sensitivity Analysis}
\vspace{-0.15cm}
We have analyzed the performance of various CL methods under budgeted computation. We have consistently observed over a variety of settings on large-scale datasets that a simple method, \ie, Naive, simply sampling with a Class-Balanced strategy and a cross entropy loss outperforms all existing methods.  However, all reported results were for $20$ time steps with $\mathcal{C} = 400$ or $\mathcal{C}=100$ training iterations for ImageNet2K and CGLM, respectively, in which expensive methods were normalized accordingly. Now, we analyze the sensitivity of our conclusions over different time steps and iterations $\mathcal{C}$.

\noindent\textbf{Does the Number of Time Steps Matter?} Prior art, such as GDumb~\cite{prabhu2020gdumb}, found that the relative performance of CL methods changes drastically when the number of time steps is varied. Subsequently, we increased the number of time steps to $50$ and $200$ from $20$, a more extreme setting than explored in recent works, while maintaining the same overall computational budget $\mathcal{C}$ eliminating any source of performance variation due to a different total computational budget. This is since per time step, the stream reveals fewer number of samples $n$ with an increased number of time steps. We report experiments in the CGLM setting where Naive will receive only $40$ and $10$ iterations for the $50$ and $200$ time steps, respectively. We consider distillation approaches where they are permitted $\nicefrac{2}{3}\mathcal{C}$, which is $27$ and $7$ iterations, respectively, on the $50$ and $200$ time steps, respectively. Note that, in these settings, per time step, methods observe $\nicefrac{2}{3} \times 11.6$K and $\nicefrac{2}{3} \times 2.9$K samples, respectively. We leave the experiments on ImageNet2K for the Appendix due to space constraints. 
We compare two distillation methods against Naive in Figure \ref{fig:difftimestepscglm}. Other methods are presented in the Appendix.

\noindent\textbf{Conclusion.} We still consistently observe that Naive outperforms all distillation methods on both the $50$ and the $200$ time steps. Moreover, the relative performance across distillation methods is preserved similarly to the $20$ time steps setup. That is, our conclusions are largely robust under different number of time steps. This is contrary to the observation of the prior art \cite{prabhu2020gdumb}, this is because unlike our setting, \cite{prabhu2020gdumb} does not scale the compute with increased number of time steps. 

\noindent\textbf{Does the Compute Budget Matter?} 
Finally, we explore the impact of changing the computational budget on the performance of different distillation methods on CGLM under $20$ time steps. We study two scenarios, one where the budget is increased to $\mathcal{C}=400$ and where it is reduced to $\mathcal{C}=40$, originally $\mathcal{C}=100$ for CGLM. Hence, distillation would be allocated $267$ and $27$ iterations in this setting, respectively. As shown in Figure \ref{fig:effective_epochs}, the higher budget setting allows approximately a full pass per time step over all stored data. We leave the experiments on ImageNet2K for the Appendix. We compare two distillation methods with Naive in Figure \ref{fig:diffstepscglm}. The remaining methods are presented in the Appendix.

\noindent \textbf{Conclusion.} Again, we observe that Naive outperforms all distillation methods in both increased and decreased compute budget settings. The final gap between MSE distillation and Naive is $11.41\%$ for $\mathcal{C}=40$, this gap is reduced to $3.85\%$ for $\mathcal{C}=400$. Surprisingly, even with increased compute budget, distillation methods still fall behind Naive. However, the reduced gap in performance compared to that of Naive is a strong indication that the reason behind the failure of distillation methods is indeed the limited computation.

 \begin{figure}
    \centering
    \includegraphics[width=0.8\linewidth]{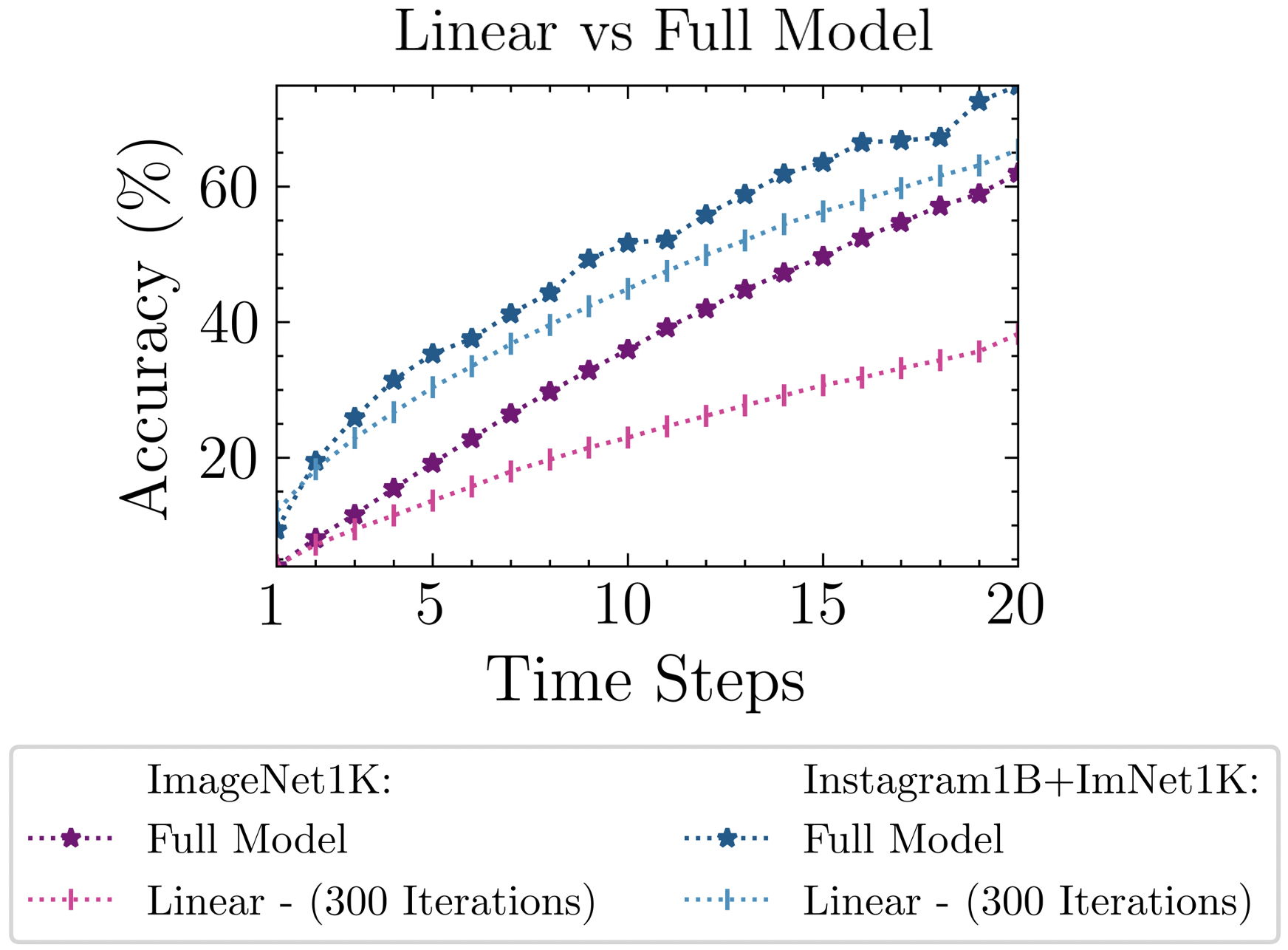}
    \vspace{-0.15cm}
    \caption{\textbf{Linear vs Full Model Training.} Performing Linear fine tuning allows to leveraging the computational budget efficiently improving the gap compared to full model training particularly for better pretrained models, \eg,  Instagram1B+ImNet1K.
    }
    \vspace{-0.6cm}
    \label{fig:lastsection}
\end{figure}
\subsection{Exploring Partial Training}
\vspace{-0.15cm}
We now investigate the reallocation of the computational budget through partial training of the model, which is a model expansion method that involves pre-selecting the subnetwork to be trained. This approach is more computationally efficient, especially on large-scale datasets. The top (FC) layer is the smallest part of the network that can be retrained. We compare partial training  of the network, \ie, FC layer only, to training the full model (Naive) using two different model initializations, ImageNet1K pretraining \cite{he2016deep} and Instagram1B+ImageNet1K pretraining \cite{mahajan2018exploring}. Note that Instagram1B+ImageNet1K is a stronger pretrained model, with better feature representations. To normalize the computation for the FC partial training, we permit $3\mathcal{C}$ training iterations compared to full model training (Naive) with $\mathcal{C}$ training iterations. Hence, CGLM FC partial training performs 300 iterations compared to 100 iterations for training Naive. We present our results in Figure \ref{fig:lastsection}, where the shades of purple and blue represent models trained from pretrained ImageNet1K and Instagram1B+ImageNet1K models, respectively.

\noindent\textbf{Conclusion.} There exists a gap between full model training and partial FC layer training (Linear). However, this gap is greatly reduced when a stronger pretrained model is adopted as an initialization. More specifically, the final gap drops from $23.73\%$ for ImageNet1K initialization to $9.45\%$ for Instagram1B+ImageNet1K initialization. Partial training of the FC layer for Instagram1B+ImageNet1K model initialization outperforms ImageNet1K full model training on average, over time steps, by $8.08\%$, which verifies that partially training a strong backbone could be more beneficial than fully training a weaker one.
\vspace{-0.15cm}
\section{Conclusion}
\label{conclusion}
\vspace{-0.15cm}
 Existing CL algorithms, such as sampling strategies, distillation, and FC layer corrections, fail in a budgeted computational setup. Simple Naive methods based on experience replay outperform all the considered CL methods. This conclusion was persistent even under various computational budgets and an increased number of time steps. We find that most CL approaches perform worse when a lower computational budget is allowed per time step.

\section{Acknowledgements}
This work was supported by the King Abdullah University of Science and Technology (KAUST) Office of Sponsored Research (OSR) under Award No. OSR-CRG2021, SDAIA-KAUST Center of Excellence in Data Science, Artificial Intelligence (SDAIA-KAUST AI), and UKRI grant: Turing AI Fellowship EP/W002981/1. We thank the Royal Academy of Engineering and FiveAI for their support. Ser-Nam Lim from Meta AI is neither supported nor has relationships with the mentioned grants. Ameya Prabhu is funded by Meta AI Grant No DFR05540.

{\small
\bibliographystyle{ieee_fullname}
\bibliography{egbib}
}

\clearpage
\appendix
\onecolumn
\section{Motivation: Privacy for Restricting Memory}

\begin{table}[h!]
\scriptsize

    \centering
    \begin{tabular}{lllcccc} \toprule
      Ref. & Dataset & Ordering & Memory & Cost & Iters & Cost \\ \midrule
      \cite{buzzega2020dark} & CIFAR10 & Cls Inc & 1-25MB & 0.05\textcent & 250K-375K & 20\$ \\
      \cite{rebuffi2017icarl}  & \multirow{2}{*}{CIFAR100} & \multirow{2}{*}{Cls Inc} & \multirow{2}{*}{10 MB} & \multirow{2}{*}{0.02\textcent} & 50K  & 8\$\\
      \cite{douillard2020podnet, hou2019learning} &  & & & & 125K  & 15\$ \\  \hline
      \cite{buzzega2020dark} & TinyImageNet & Cls Inc & 5-20 MB & 0.04\textcent & 350K-500K & 25\$\\ 
      \cite{hou2019learning, douillard2020podnet} & ImageNet100 & Cls Inc & 0.3-1 GB & 2\textcent & 100K & 50\$ \\ 
      \cite{hou2019learning, douillard2020podnet} & ImageNet1K & Cls Inc & 33GB & 66\textcent & 1M & 500\$ \\
      \cite{lin2021clear} & CLEAR & Dist Shift & 0.4-1.2GB & 2\textcent & 300K & 100\$ \\\midrule        
      \cite{he2016deep} & ResNet50 (bs=256) & & 22GB & & \\ \midrule
      \multirow{2}{*}{Ours} & GLDv2-CL & Dist Shift & \textbf{90GB} & 2\$ & 2K  & 10\$\\
      & ImageNet21K-CL & ClsInc, DataInc & \textbf{400GB} & 10\$ & 8K  & 35\$\\ \bottomrule
    \end{tabular}
    \vspace{-0.2cm}
    \caption{\textbf{Cost of Memory vs Computation.}  Google Cloud Standard Storage (2\textcent  per  GB per month for 1 month) and Compute Cost measured as running cost of an A2 instance (3\$ per hour for 1 GPU). Number of iterations (forward+backward passes) for training a CL model on that dataset listed for  comparison invariant to input image and model sizes. We observe that computational costs for running an experiment far outweigh the costs for storing replay samples.}
    \vspace{-0.35cm}
    \label{table:cost_intro}
\end{table} 

Prior art on continual learning \cite{lopez2017gradient,rebuffi2017icarl,chaudhry2018efficient,buzzega2020dark,aljundi2019gradient,lin2021clear} motivate the problem from the aspect of prohibited access to previously received data; except for a small portion that is allowed to store in memory. The two principal motivations behind restricting the access to past samples in the literature are two folds.
(\textbf{i}) Storage space is expensive. (\textbf{ii}) Access to previous data is prohibitive due to privacy and GDPR constraints. 

\textbf{Cost.} As for the first argument used as a motivation for limiting the memory size, as we have elaborated in Section 1 of the main paper and similarly further detail in Table \ref{table:cost_intro}, the cost of storing data is insignificant. This is particularly the case when considering the associated computational costs of training deep models. For example, as per Table \ref{table:cost_intro}, it costs 2 cents to store the entirety of the CLEAR dataset, among the largest datasets for continual learning, while it costs about $100\$$ to train a model continually on the same dataset.
If reducing costs are the key issue, limited computational budget and not memory, as argued earlier, is the way forward.

\textbf{Privacy.} A classical argument is that due to GDPR requirements, data needs to be removed or company privacy policies, it can no longer accessible after certain period, typically a few months. Any previous benchmark with memory constraints already violates this privacy consideration. This is since, which data is to be made private and shall be deleted should not be up to the learning algorithm to decide. But, we further argue that simply not storing any samples still does not address privacy considerations. Deep models store information about trained samples \cite{jagielski2020high, jagielski2022measuring}. A glaring violation additionally was presented in Haim \etal \cite{haim2022reconstructing}, showing that we can reconstruct a large number of training samples only given the trained model. Without removing the information explicitly from the deep model, allowing no access to past samples does not alleviate privacy concerns. Goel \etal  \cite{goel2022evaluating} presents a catastrophic forgetting baseline for inexact machine unlearning,  indicating that forgetting might be the very objective of sustaining privacy, antithetical to the objective of continual learning.

\section{Dataset Construction}

\subsection{Constructing Imagenet2K}

ImageNet2K train set is constructed using all training images in ImageNet1K dataset \cite{deng2009imagenet} for 1K classes as an initialization, with selecting an additional 1K non-overlapping classes from ImageNet21K dataset \cite{deng2009imagenet} to form the ImageNet2K dataset. We illustrate the creation of the test, validation and train sets below:

\textbf{Test Set}: We use the ImageNet1K val set as the test set to be  consistent with test sets used in previous literature using ImageNet1K. We separate 50 images per class from the sample set of the new 1K classes. We combine these two sets to create the overall test set for experiments. The test set for every timestep consists of classes from this test set which have been seen so far.

\textbf{Validation Set}: We use ImagenetV2 dataset \cite{recht2019imagenet} as the validation set from Imagenet1K data. We seperate 50 images per class, not used in the test set to create the val set for the new 1K classes. We combine these two sets to create the overall validation set for experiments. The validation set for every timestep consists of classes from this validation set which have been seen so far.

\textbf{Train Set}: We order all the samples from the new 1K classes not used for creating the test and val sets for training. We order them by classes to form the CI-ImageNet2K stream and randomly shuffle all these images to form the DI-ImageNet2K stream. Note that the stream order is provided samplewise, allowing the stream size $N$ to be adjusted. In the standard experiments, data equivalent to 50 classes is sampled every timestep, for 20 timesteps.


\subsection{Constructing Continual Google Landmarks V2}

Continual Google Landmarks V2 (CGLM) consists of 580K samples. To obtain this subset, we start with the train-clean subset of the Google Landmarks V2 available from the Google Landmarks V2 dataset website\footnote{https://github.com/cvdfoundation/google-landmark}. We apply the following preprocessing steps in order:
\begin{enumerate}
    \setlength\itemsep{-0.25em}
    \item Filter out images which do not have timestamp metadata available. 
    \item Remove images of classes that have less than 25 samples in total
    \item Order data by timestamp.
    \item Randomly sample 10\% of data from across time as the test set
\end{enumerate}
 We get the rest $580K$ images as the train set for continual learning over $10788$ classe, with rapid temporal distribution shifts. We do not have a validation set here as we benchmark transfer of hyperparameters used from ImageNet to this dataset.

\section{Estimation of Equivalent Iterations}

In this section, we elaborate on the details of selecting the computational budget for distillation and expensive sampling approaches. The key point for these calculations is the fact that the computational (and time) cost for a forward pass is $\nicefrac{1}{2}$ the cost of a backward pass \footnote{https://www.lesswrong.com/posts/jJApGWG95495pYM7C/how-to-measure-flop-s-for-neural-networks-empirically}. 

\textbf{Distillation}: When distillation approaches have a budget of $\nicefrac{2}{3}^{rd}$ iterations of the naive baseline, the computational cost is as follows: $\nicefrac{2}{3}\mathcal{C}$ for training of the student model, and $\nicefrac{1}{2}\times \nicefrac{2}{3} = \nicefrac{1}{3} \mathcal{C}$ for the teacher model which only has a forward pass, which sums up to $\mathcal{C}$. Hence, distillation methods have an  equivalent computational budget as the naive baseline with $\nicefrac{2}{3}^{rd}$ training iterations.

\textbf{Sampling}: We train the expensive models for $\nicefrac{1}{2}$ the number of iterations as a naive model. To select that subset of training data, we randomly sample $3\times$ the required number of training samples from the stored set and forward pass them through the latest trained model to obtain the features/probabilities. And then we select the best $\nicefrac{1}{3}^{rd} $ of the $3\times$ set for training using different selection functions. We assume the cost of selecting samples given the features/probabilities is negligible.

The computational cost of training for expensive sampling methods is $\nicefrac{1}{2}\mathcal{C}$, as the selected sample set is half the size compared to the naive baseline. The computational cost of selecting the samples is $\nicefrac{1}{2} \times 3 \times \nicefrac{1}{3} \mathcal{C} = \nicefrac{1}{2}\mathcal{C}$ (forward pass requires $\nicefrac{1}{3}^{rd}$ of the total cost, on $3\times$ the required data, the required data size being $\nicefrac{1}{2}$ when compared to naive). The combined cost is the sum of selection and training cost, which is $\mathcal{C}$. Hence, expensive sampling methods have an equivalent budget with $\nicefrac{1}{2}$ training iterations.

\section{Additional Results}

\begin{figure*}[htbp!]
    \centering
    \includegraphics[width=\textwidth]{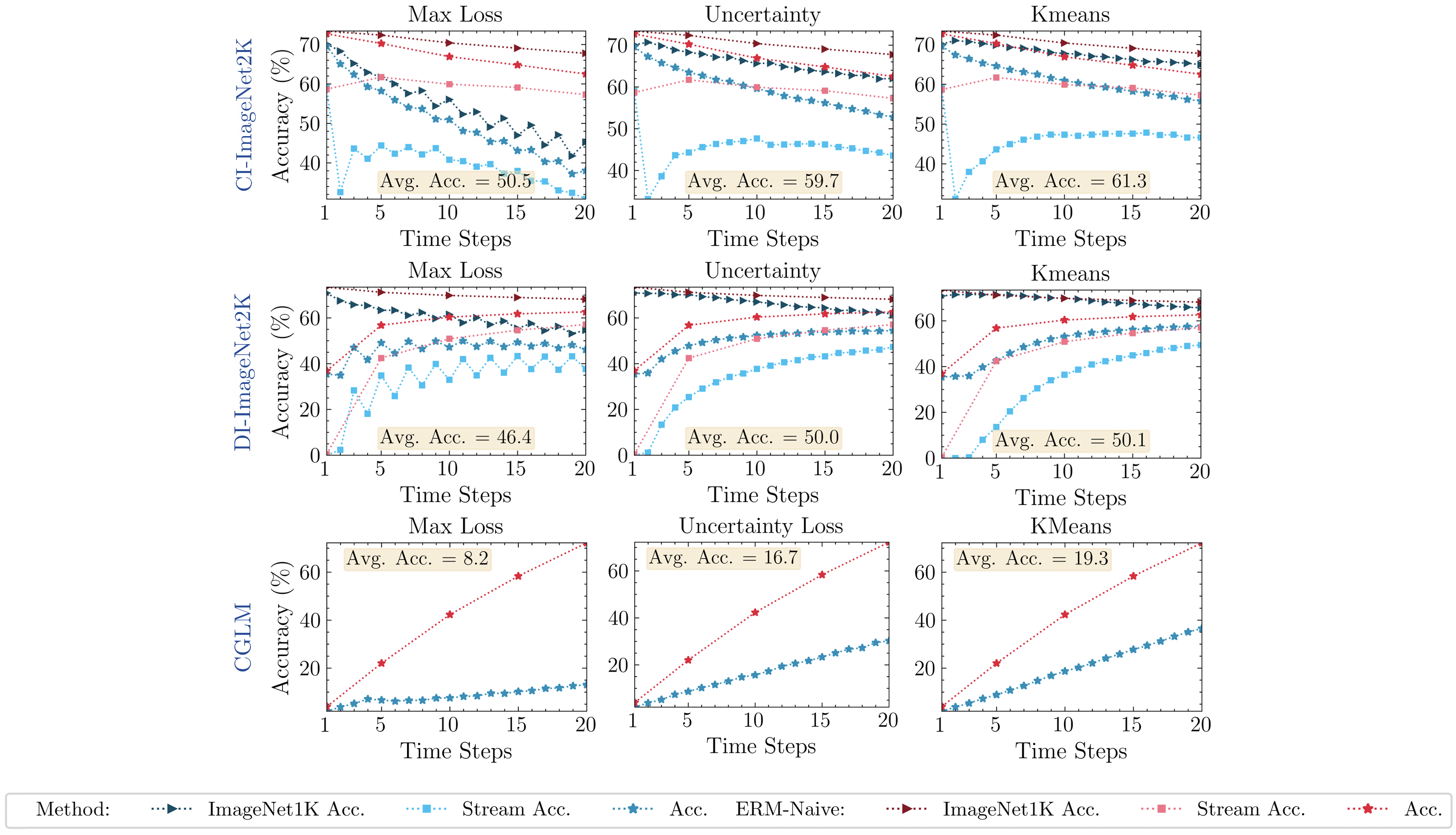}
    \caption{\textbf{Expensive Sampling.} As mentioned in the manuscript, KMeans performs the best among expensive sampling techniques such as Max Loss and Uncertainty Loss. Nevertheless, the performance of the expensive sampling methods is worse than simple Naive.}
    \label{fig:expensive_sampling}
\end{figure*}

\begin{figure*}[htbp!]
    \centering
    \includegraphics[width=\textwidth]{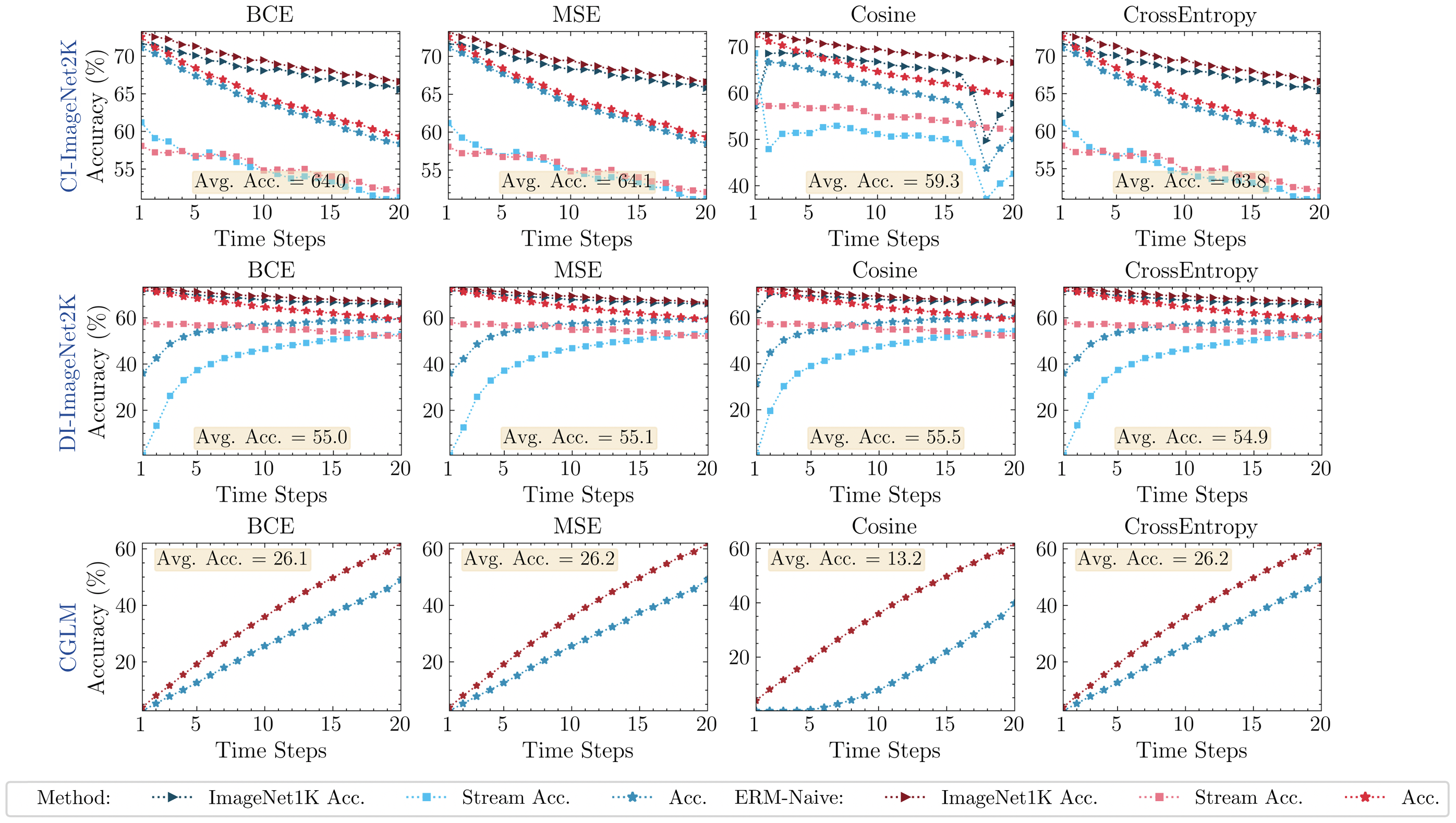}
    \caption{\textbf{Distillation Methods.} All four studied distillation methods under perform compared to the simple Naive baseline in all three studied settings. ImageNet experiments are allowed 400 iterations whereas CGLM is allowed 100 iterations.}
    \label{fig:distillation_main}
\end{figure*}

Due to limited space, some of the experiments in the manuscripts were deferred to the Appendix. In this section we present results for all mentioned costly sampling methods and distillation methods. Additional results for   
\textbf{\textit{Section 4.2: 1 ``Do Sampling Strategies Matter?"}} are presented in Figure \ref{fig:expensive_sampling} where all three costly sampling methods are presented, namely Max Loss, Uncertainty Loss, and KMeans. Similarly, additional results for \textbf{\textit{Section 4.2: 2 ``Does Distillation Matter?"}} are presented in Figure \ref{fig:distillation_main} where all four distillation methods are presented, namely BCE, MSE, Cosine, and CrossEntropy. We observe that all previous conclusions consistently hold, \ie the Naive baseline is still leading in comparison to all previous approaches.

\begin{figure}[htbp!]
    \centering
\includegraphics[width=\textwidth]{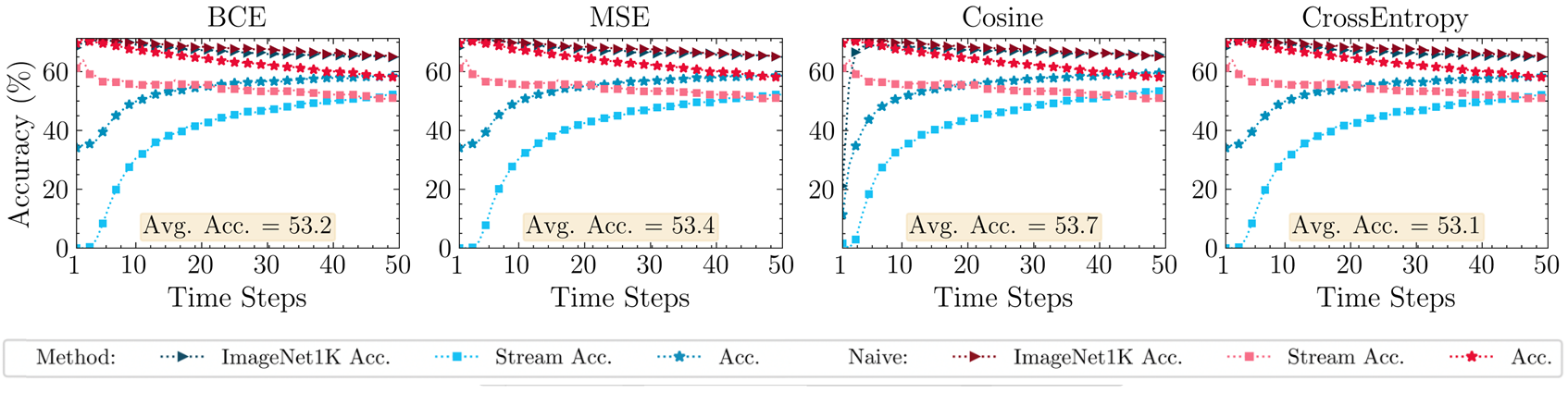}
    \caption{\textbf{DI-ImageNet2K 50 Time Steps.} As observed in the manuscript, when the number of time steps, distillation methods still under perform compared to the Naive baseline.}
    \label{DIImageNet50}
\end{figure}

\begin{figure}[htbp!]
    \centering
\includegraphics[width=\textwidth]{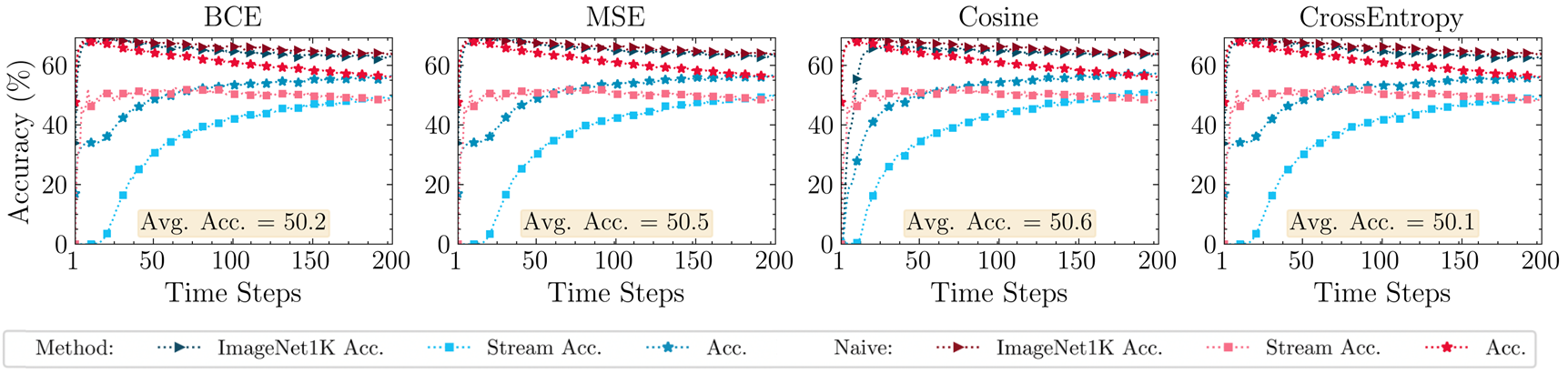}
    \caption{\textbf{DI-ImageNet2K 200 Time Steps.} As observed in the manuscript, when the number of time steps, distillation methods still under perform compared to the Naive baseline.}
    \label{DIImageNet200}
\end{figure}

\begin{figure}[htbp!]
    \centering
\includegraphics[width=\textwidth]{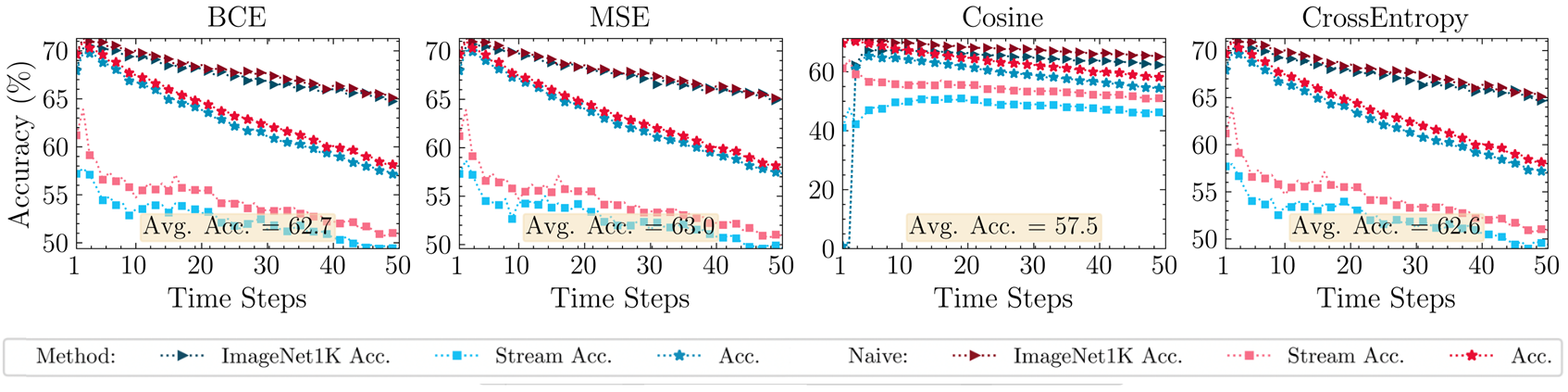}
    \caption{\textbf{CI-ImageNet2K 50 Time Steps.} As observed in the manuscript, when the number of time steps, distillation methods still under perform compared to the Naive baseline.}
    \label{CIImageNet50}
\end{figure}

\begin{figure}[htbp!]
    \centering
\includegraphics[width=\textwidth]{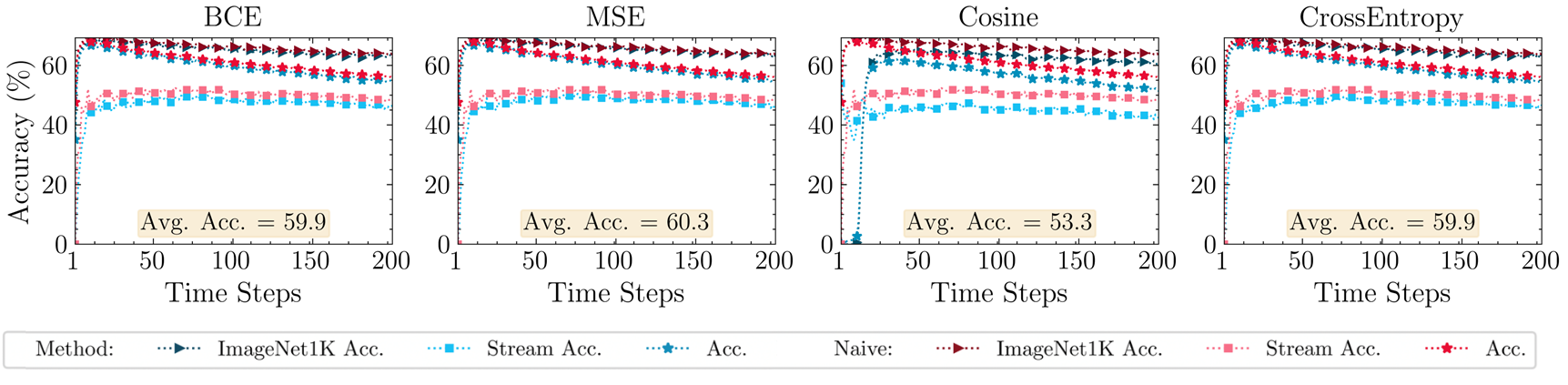}
    \caption{\textbf{CI-ImageNet2K 200 Time Steps.} As observed in the manuscript, when the number of time steps, distillation methods still under perform compared to the Naive baseline.}
    \label{CIImageNet200}
\end{figure}

\begin{figure}[htbp!]
    \centering
\includegraphics[width=\textwidth]{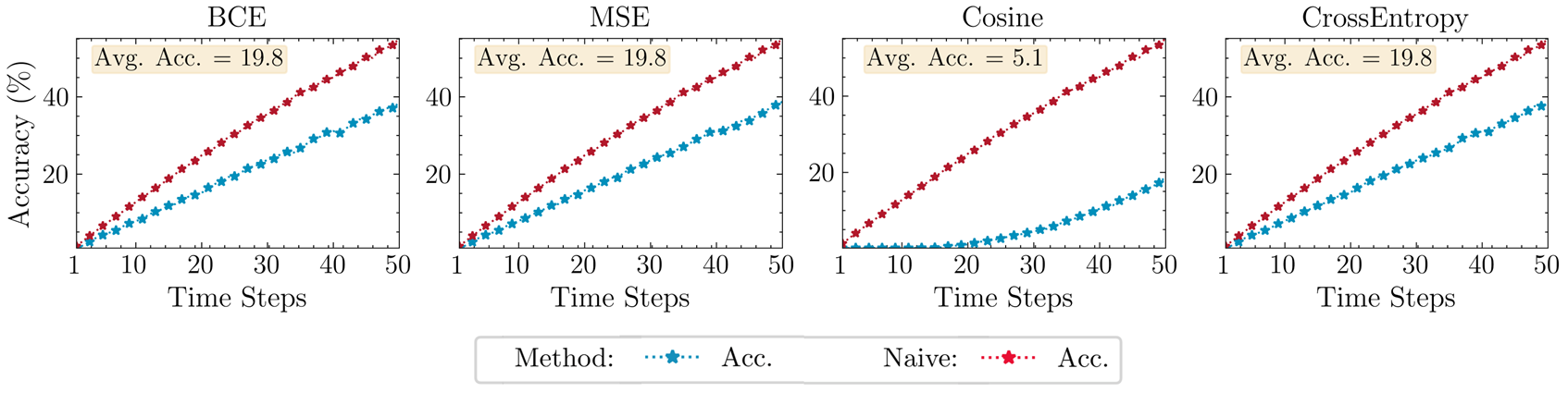}
    \caption{\textbf{CLGM 50 Time Steps.} As observed in the manuscript, when the number of time steps, distillation methods still under perform compared to the Naive baseline.}
    \label{CGLM50}
\end{figure}

\begin{figure}[htbp!]
    \centering
\includegraphics[width=\textwidth]{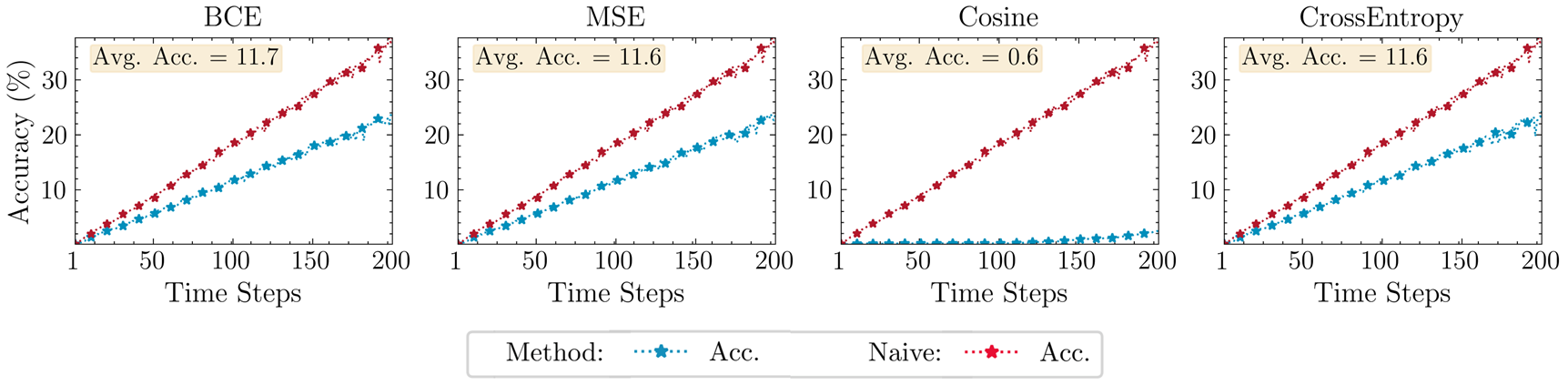}
    \caption{\textbf{CGLM 200 Time Steps.} As observed in the manuscript, when the number of time steps, distillation methods still under perform compared to the Naive baseline.}
    \label{CGLM200}
\end{figure}

We also extend the time steps and the number of iterations sensitivity experiments to all four considered distillation methods in all three setups, DI-ImageNet2K, CI-ImageNet2K and CGLM. We present additional results for \textit{\textbf{Section 4.3: 1. Does the Number of Time Steps Matter?}}: with results for DI-ImageNet2K presented in Figures \ref{DIImageNet50} and \ref{DIImageNet200} with 50 and 200 time steps respectively, CI-ImageNet2K in Figure \ref{CIImageNet50} and \ref{CIImageNet200} with 50 and 200 time steps respectively and CGLM in Figures \ref{CGLM50} and \ref{CGLM200}  with 50 and 200 time steps respectively. We observe that all previous conclusions consistently hold, \ie the conclusions are robust to changing time steps for a given cost $\mathcal{C}$.

\begin{figure}[htbp!]
    \centering
\includegraphics[width=\textwidth]{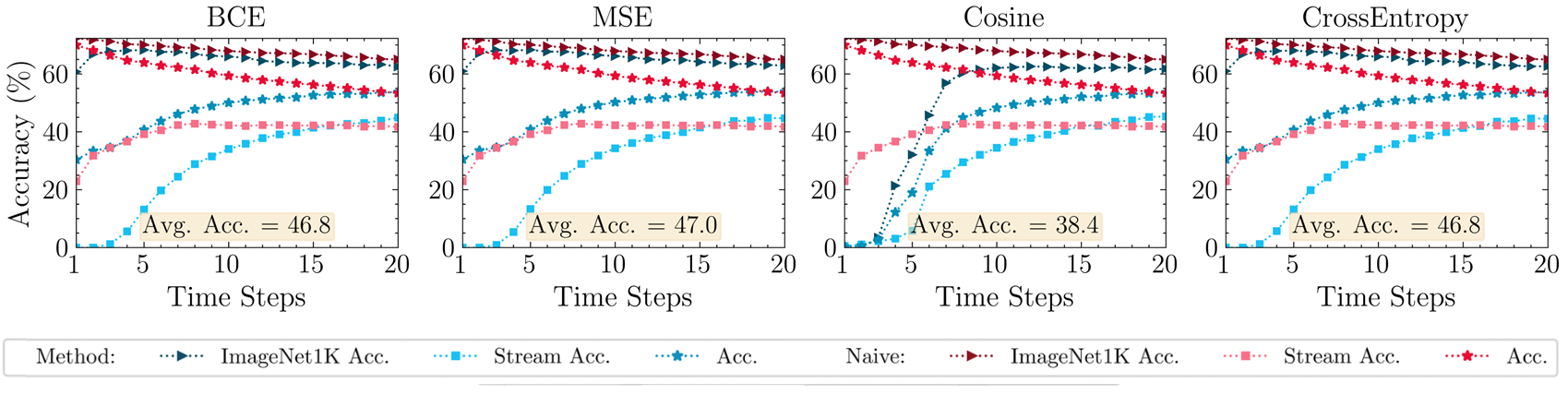}
    \caption{\textbf{DI-ImageNet2K - 100 Iterations.} As observed in the manuscript, with reduced compute, distillation methods still under perform compared to the Naive baseline. The compute budget of the Naive baseline, $\mathcal{C}$, is set to 100 iterations whereas that of the distillation methods is $\nicefrac{2}{3}~\mathcal{C} = 67$ iterations.}
    \label{DIImageNet100}
\end{figure}

\begin{figure}[htbp!]
    \centering
\includegraphics[width=\textwidth]{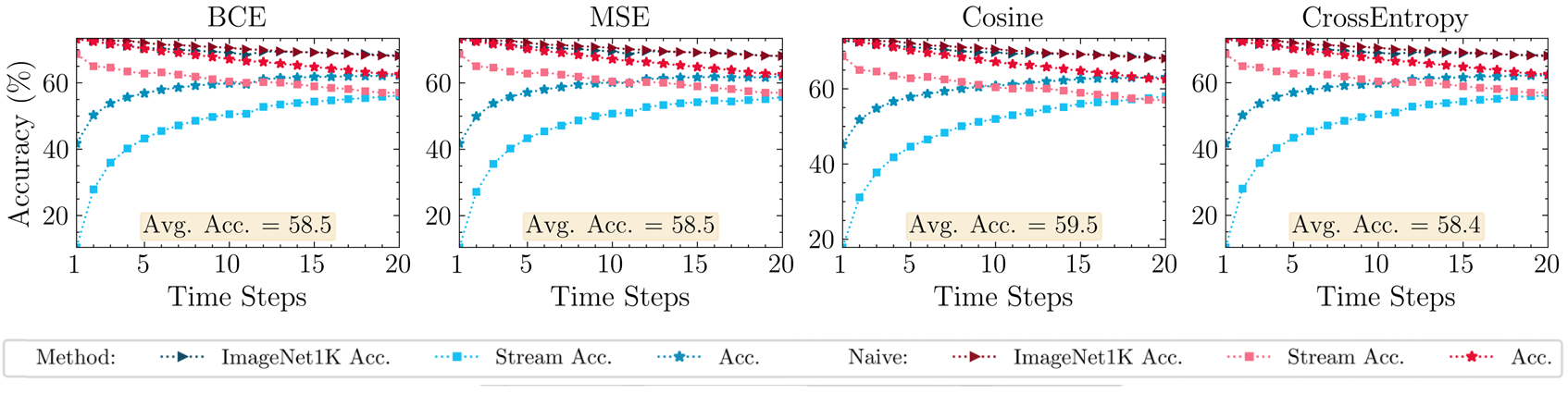}
    \caption{\textbf{DI-ImageNet2K - 1200 Iterations.} As observed in the manuscript, with increased compute, distillation methods still under perform compared to the Naive baseline. The compute budget of the Naive baseline, $\mathcal{C}$, is set to 1200 iterations whereas that of the distillation methods is $\nicefrac{2}{3}~\mathcal{C} = 800$ iterations.}
    \label{DIImageNet1200}
\end{figure}

\begin{figure}[htbp!]
    \centering
\includegraphics[width=\textwidth]{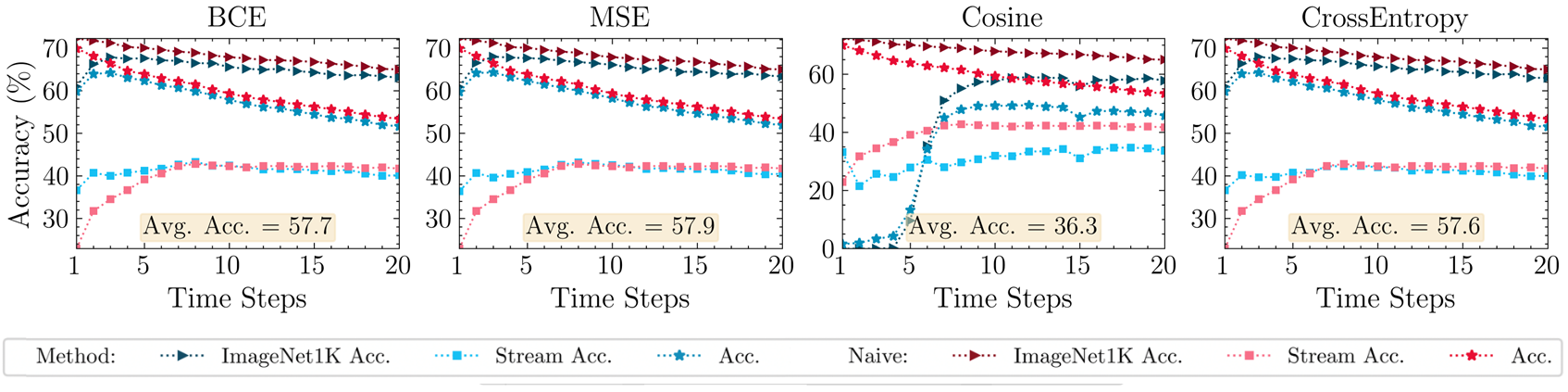}
    \caption{\textbf{CI-ImageNet2K - 100 Iterations.} As observed in the manuscript, with reduced compute, distillation methods still under perform compared to the Naive baseline. The compute budget of the Naive baseline, $\mathcal{C}$, is set to 100 iterations whereas that of the distillation methods is $\nicefrac{2}{3}~\mathcal{C} = 67$ iterations.}
    \label{CIImageNet100}
\end{figure}

\begin{figure}[htbp!]
    \centering
\includegraphics[width=\textwidth]{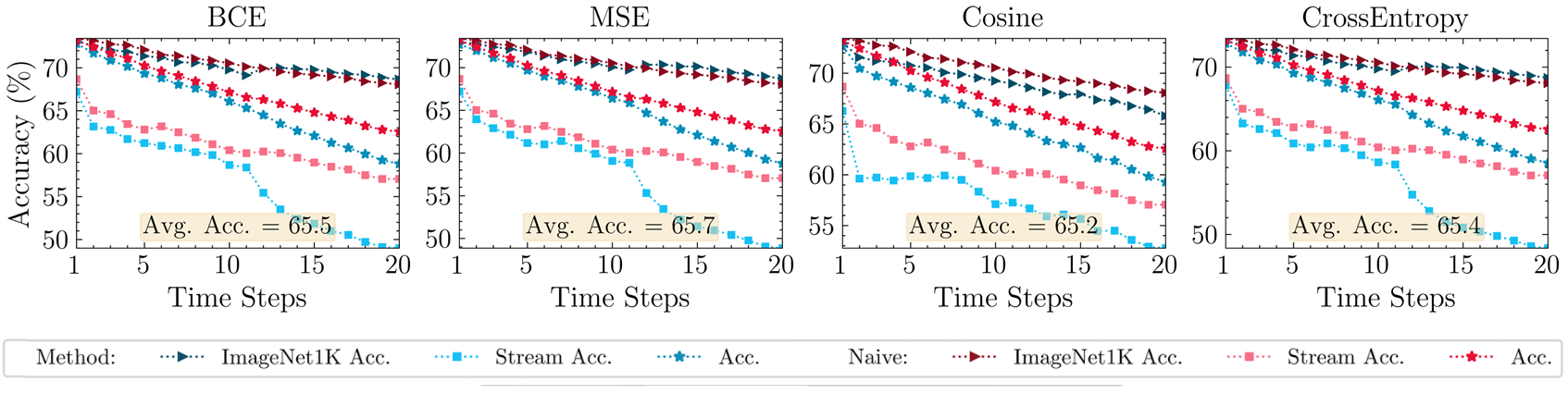}
    \caption{\textbf{CI-ImageNet2K - 1200 Iterations.} As observed in the manuscript, with increased compute, distillation methods still under perform compared to the Naive baseline. The compute budget of the Naive baseline, $\mathcal{C}$, is set to 1200 iterations whereas that of the distillation methods is $\nicefrac{2}{3}~\mathcal{C} = 800$ iterations.}
    \label{CIImageNet1200}
\end{figure}

\begin{figure}[htbp!]
    \centering
\includegraphics[width=\textwidth]{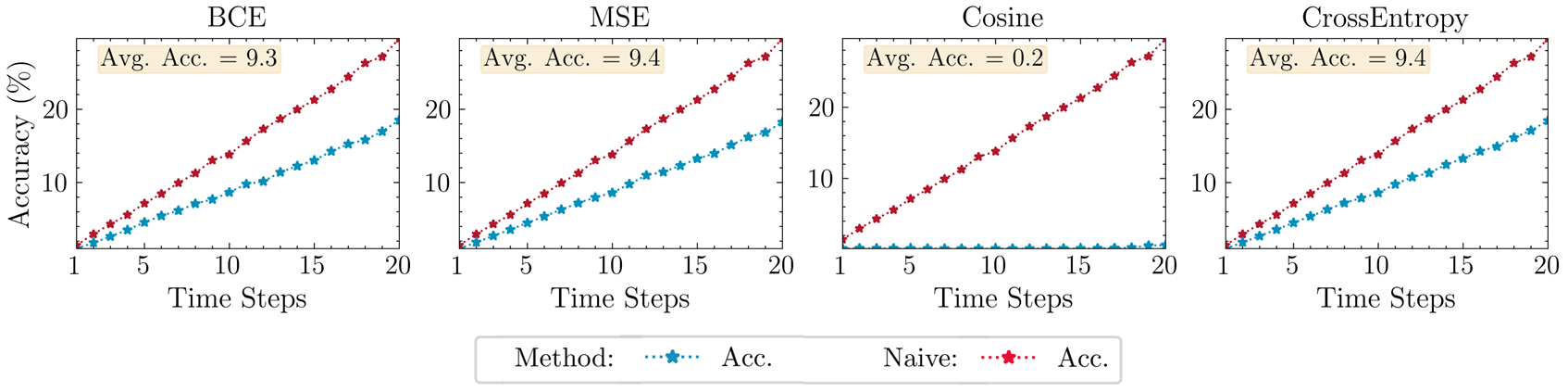}
    \caption{\textbf{CGLM 40 - Iterations.} As observed in the manuscript, with reduced compute, distillation methods still under perform compared to the Naive baseline. The compute budget of the Naive baseline, $\mathcal{C}$, is set to 40 iterations whereas that of the distillation methods is $\nicefrac{2}{3}~\mathcal{C} = 27$ iterations.}
    \label{CGLM40}
\end{figure}

\begin{figure}[htbp!]
    \centering
\includegraphics[width=\textwidth]{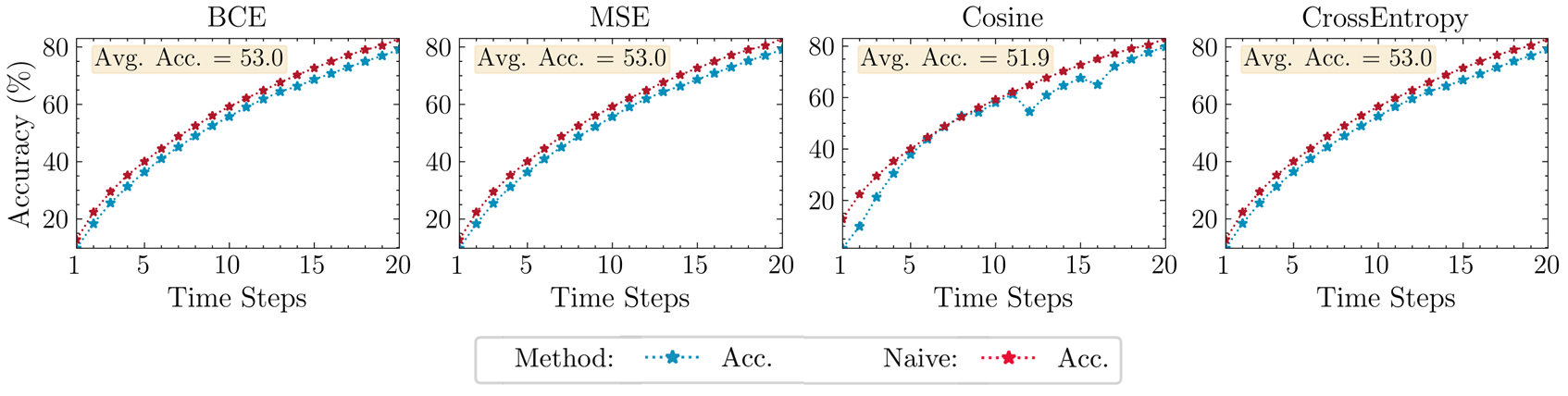}
    \caption{\textbf{CGLM 400 - Iterations.} As observed in the manuscript, with increased compute, distillation methods still under perform compared to the Naive baseline. The compute budget of the Naive baseline, $\mathcal{C}$, is set to 400 iterations whereas that of the distillation methods is $\nicefrac{2}{3}~\mathcal{C} = 267$ iterations.}
    \label{CGLM400}
\end{figure}

We present additional results for \textit{\textbf{Section 4.3: 2. Does the Compute Budget Matter?}}: on DI-ImageNet2K in Figures \ref{DIImageNet100} and \ref{DIImageNet1200} for 100 and 1200 iterations respectively, and CI-ImageNet2K in Figures \ref{CIImageNet100} and \ref{CIImageNet1200} for 100 and 1200 iterations respectively and on CGLM in Figures \ref{CGLM40} and \ref{CGLM400} for 40 and 400 iterations respectively. We observe that all previous conclusions consistently hold, \ie the conclusions are robust to changing computational cost, towards both harsher and laxer computational constraint regimes.

\clearpage

\subsection{Effect of Weight Decay}

\begin{figure}[h!]
    \centering
    \includegraphics[width=0.8\linewidth]{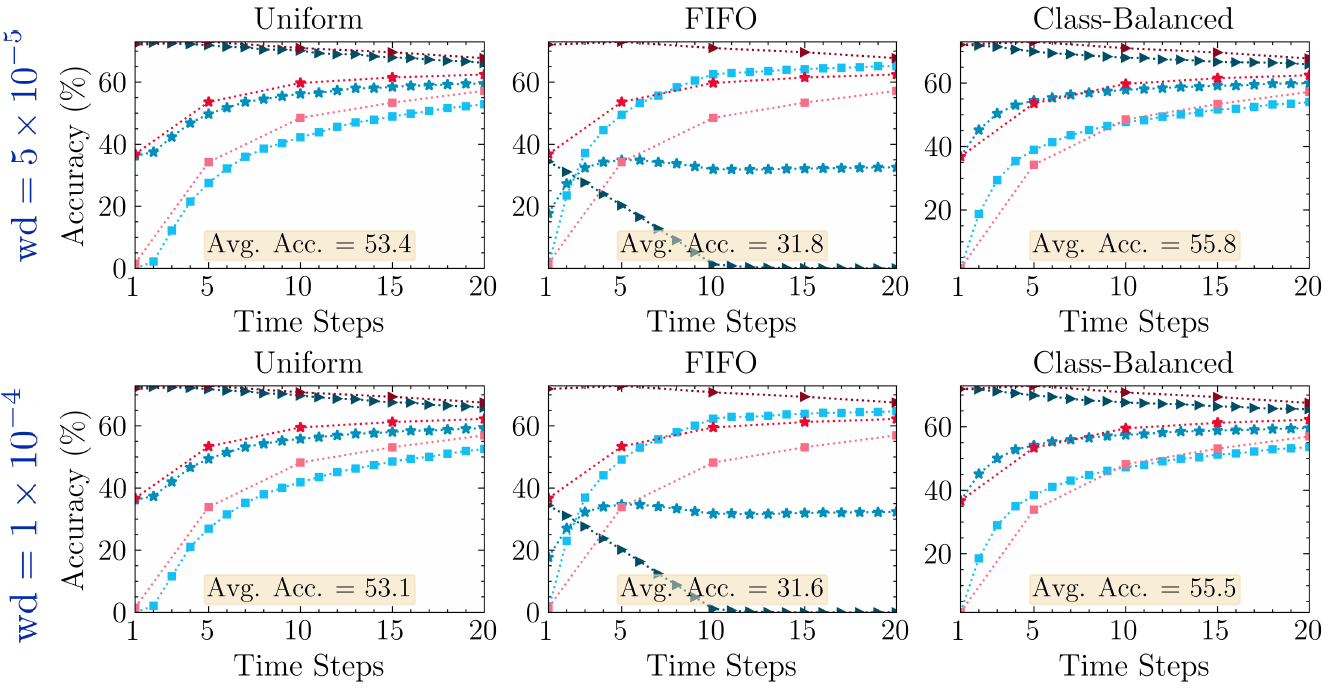}
    \caption{\textbf{Effect of Weight Decay.} Increasing the weight decay causes a slight drop in performance. Setting wd=$0$ gave the best results during our parameter cross-validation.}
    \label{fig:wd_ablation}
\end{figure}

The choice of weight decay, $\text{wd}=0$, in the manuscript was based on  result of cross-validation from the set. More specifically, we try weight decays of $\{5\times 10^{-5}, 1\times 10^{-4}\}$. We observe a minor difference in performance between various weight decays, with a wd=$0$ consistently being slightly better. The results are shown in Figure \ref{fig:wd_ablation}

\subsection{Effect of Batch Size}
\begin{figure}[h!]
    \centering
    \includegraphics[width=0.7\textwidth]{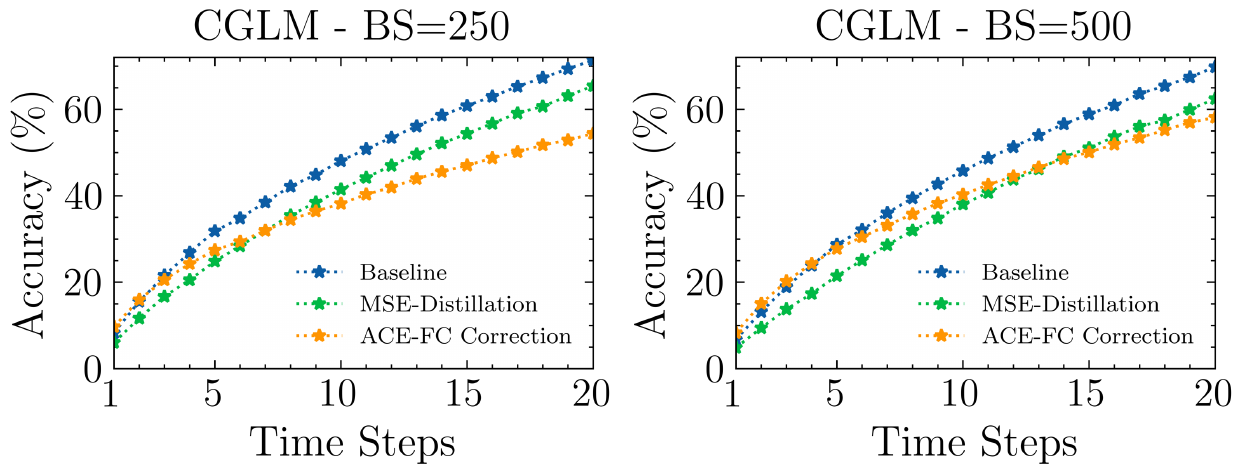}
    \caption{\textbf{Effect of Batch Size.} The conclusions presented in our work hold even when the batch size while is changed while the same overall computational budget.}
    \label{fig:bs_ablation}
\end{figure}

In all experiments, we fixed the batch size (BS) to 1500 to optimize the utilization of our hardware resources and minimize the training time. As shown in the literature, BS and learning rate (LR) are closely related. The selected LR was tuned to fit the selected BS. Regardless, we present varying BS experiments in Fig (1)
where we study the latest FC correction method, ACE, and the distillation method, MSE, for BS of 250 and 500 with  increased iterations of 600 and 300, respectively, and crossvalidated learning rates. We observe that the Naive baseline
is still superior even when the batch size is adjusted. Our findings are summarized in Figure \ref{fig:bs_ablation}.

\subsection{Effect of Increasing Computational Budget on Distillation}

\begin{figure}[h!]
    \centering
    \includegraphics[width=0.7\linewidth]{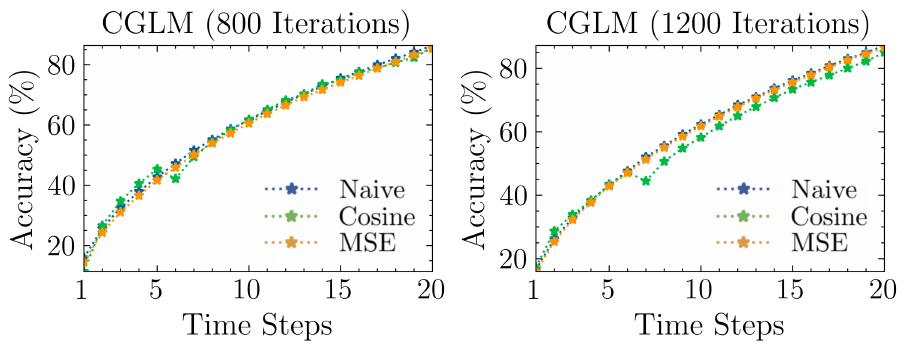}
    \caption{\textbf{Effect of Increasing Computation Budget on Distillation.} As the computation budget is increased while maintaining a normalized compute among different methods, Naive baseline still outperforms distillation based methods.}
    \label{fig:distill_steps_rebuttal}
\end{figure}

We complement the results in Figure \ref{fig:diffstepscglm} with additional experiments using $800$ and $1200$ iterations.  The observed results align with our previous observations; as long as the computation is normalized across methods, naive, the most simplest among the considered methods, outperforms existing methods. This is as opposed to prior art comparison that does not normalize compute, which puts the Naive baseline in disadvantage. The results are shown in Figure \ref{fig:distill_steps_rebuttal}.

\end{document}